\pdfoutput=1

\documentclass[11pt]{article}
\usepackage{algpseudocode}
\usepackage[final]{acl}

\usepackage{times}
\usepackage{latexsym}
\usepackage{amsmath}
\usepackage{amssymb}
\usepackage{amsthm}
\newtheorem{proposition}{Proposition}
\usepackage{soul}
\usepackage{graphicx}
\usepackage{algorithm}
\usepackage{textcomp}
\usepackage{tikz}
\usepackage[dvipsnames]{xcolor}
\usepackage{pgfplots}

\usepackage{newunicodechar}
\newunicodechar{✓}{\checkmark}
\pgfplotsset{compat=1.17}
\usepackage{colortbl}
\usepackage{pifont}
\usepackage{stfloats}
\newcommand{\cmark}{\ding{51}}
\newcommand{\xmark}{\ding{55}}
\usepackage{cuted}

\usepackage{makecell} 
\usepackage{booktabs}
\usepackage{multicol,multirow}
\usepackage{adjustbox}
 \usepackage{colortbl}
\definecolor{bestresult}{RGB}{220,255,220} 
\definecolor{secondbest}{RGB}{220,220,255} 
\usepackage{tabularx}
\usepackage{float}
\algrenewcommand{\algorithmicrequire}{\textbf{Require:}}
\usepackage{pifont}

\usepackage{enumitem}
\usepackage{todonotes}
\usepackage{subcaption}

\definecolor{apricot}{RGB}{251,206,177} 

\definecolor{trustgreen}{RGB}{200, 230, 200}
\definecolor{datablue}{RGB}{200, 215, 235}
\definecolor{reviewyellow}{RGB}{255, 245, 200}
\definecolor{abstainred}{RGB}{245, 200, 200}

\newcommand{\h}{\mathbf{h}}                    
\newcommand{\cls}{f_{\text{cls}}}              

\newcommand{\ck}{c_k}                          
\newcommand{\cvec}{\mathbf{c}}                 
\newcommand{\phat}{\hat{p}}                    

\newcommand{\credal}{\mathcal{C}}              
\newcommand{\plower}{\underline{p}}            
\newcommand{\pupper}{\overline{p}}             

\definecolor{trustgreen}{RGB}{198, 239, 206}      
\definecolor{dataorange}{RGB}{255, 235, 156}      
\definecolor{reviewblue}{RGB}{189, 215, 238}      
\definecolor{escalatered}{RGB}{255, 199, 206}     
\definecolor{generaluq}{RGB}{245, 245, 245}    
\definecolor{cbmrows}{RGB}{255, 250, 240}      
\definecolor{encoderrows}{RGB}{230, 242, 255}     
\definecolor{llmrows}{RGB}{232, 245, 233}      

\usepackage[T1]{fontenc}

\usepackage[utf8]{inputenc}
\usepackage{soul}

\usepackage{microtype}

\usepackage{inconsolata}
\usepackage{graphicx}
\newcommand{\modelname}{CREDENCE}

\title{Credal Concept Bottleneck Models for Epistemic–Aleatoric \\Uncertainty Decomposition}

\author{
Tanmoy Mukherjee\textsuperscript{1}, 
Thomas Bailleux,\textsuperscript{1}, 
Pierre Marquis\textsuperscript{1,2}, 
Zied Bouraoui\textsuperscript{1} 
\\
\textsuperscript{1} Univ. Artois, CNRS, CRIL, France,  
\textsuperscript{2} Institut Universitaire de France
\\
\texttt{\{mukherjee,bailleux,marquis,bouraoui\}@cril.fr}
}

\begin{document}
\maketitle

\begin{abstract}
Concept Bottleneck Models (CBMs) predict through human-interpretable concepts, but they typically output point concept probabilities that conflate epistemic uncertainty (reducible model underspecification) with aleatoric uncertainty (irreducible input ambiguity). This makes concept-level uncertainty hard to interpret and, more importantly, hard to act upon. We introduce \modelname{} (Credal Ensemble Concept Estimation), a CBM framework that decomposes concept uncertainty by construction. \modelname{} represents each concept as a credal prediction (a probability interval), derives epistemic uncertainty from disagreement across diverse concept heads, and estimates aleatoric uncertainty via a dedicated ambiguity output trained to match annotator disagreement when available. The resulting signals support prescriptive decisions: automate low-uncertainty cases, prioritize data collection for high-epistemic cases, route high-aleatoric cases to human review, and abstain when both are high. Across several tasks, we show that epistemic uncertainty is positively associated with prediction errors, whereas aleatoric uncertainty closely tracks annotator disagreement, providing guidance beyond error correlation. Our implementation is available at the following link: \url{https://github.com/Tankiit/Credal_Sets/tree/ensemble-credal-cbm}
\end{abstract}

\section{Introduction}
\label{sec:introduction}
In high-stakes settings, such as healthcare, legal analysis, and financial advising, NLP systems are increasingly used to support consequential decisions. In these contexts, it is not enough to output a prediction; practitioners must also assess when the system is reliable and, crucially, \emph{why} it may be unreliable. A key distinction lies between \emph{epistemic uncertainty} (what the model has not yet learned but could learn with additional evidence) and \emph{aleatoric uncertainty} (irreducible ambiguity or noise in the input) \citep{kendall2017uncertainties,shaker2020aleatoric}. Existing models face challenges when dealing with uncertainty \citep{hullermeier2022quantification,shaker2020aleatoric}, as they generate a single confidence score that merges various types of uncertainty, leading to a uniform approach for managing all uncertain predictions.
Consider a restaurant review classifier that processes the sentence: \emph{``The waiter was efficient, I suppose.''} A model might assign a $60\%$ probability that the sentiment about service is positive. However, this single number conflates two qualitatively different situations. The model might be uncertain because it has seen few hedged expressions such as ``I suppose'' during training (epistemic uncertainty). Alternatively, the text may genuinely express ambivalence, so that competent annotators could disagree even with unlimited training data (aleatoric uncertainty). The distinction is actionable: high epistemic uncertainty suggests targeted data collection or modeling changes, while high aleatoric uncertainty suggests human review, user-facing ambiguity communication, or abstention.

Making uncertainty actionable often requires exposing \emph{what} the model is uncertain about, not only \emph{how much} uncertainty it has. For many NLP decisions, practitioners reason in terms of interpretable attributes—e.g., whether a review expresses praise for \emph{service}, whether a sentence contains \emph{hedging}, or whether a post includes \emph{insults}. This suggests targeting uncertainty at the level of such intermediate attributes. Concept Bottleneck Models (CBMs) \cite{koh2020concept}  provide a natural framework for this setting because they explicitly use a set of human-interpretable concepts before producing the final label. However, standard CBMs typically represent each concept as a point probability, which collapses epistemic and aleatoric effects at the concept layer. As a result, concept-level uncertainty becomes difficult to interpret and, more importantly, difficult to act upon: it may indicate which predictions are risky, but not what response is appropriate.

Different uncertainty types should trigger different interventions: high epistemic uncertainty calls for collecting additional evidence or improving the model; high aleatoric uncertainty calls for human oversight or qualified outputs; and high–high cases motivate abstention or escalation. This issue is particularly pronounced in NLP: tasks involving sentiment, toxicity, and emotion are fundamentally subjective, i.e.\ “ground truth” reflects a composite of disputed human judgments. As a result, distinguishing between epistemic and aleatoric uncertainty is not just a theoretical refinement but a practical requirement for any triage system deployed in production.
Without decomposition, uncertain cases are treated uniformly, conflating fixable model limitations with irreducible ambiguity.
To address this, we introduce \modelname{} (Credal Ensemble Concept Estimation), a CBM framework that decomposes concept-level uncertainty by construction. The key insight is \textit{structural separation}: epistemic and aleatoric signals must come from different sources to prevent them from collapsing. Concretely, we compute epistemic uncertainty from \textit{disagreement across} ensemble heads and aleatoric uncertainty from a dedicated head trained \textit{within} each example on annotator disagreement.

Our contributions are: (i) a credal CBM formulation that represents concept predictions as probability intervals and supports decisions (trust, abstain, escalate to human); (ii) a structurally separated decomposition of concept uncertainty into epistemic (ensemble disagreement) and aleatoric (predicted ambiguity) components derived from different parameters and avoiding collapse; and (iii) empirical evaluation across several NLP tasks, showing that epistemic uncertainty aligns with prediction errors, whereas aleatoric uncertainty tracks annotator disagreement, yielding actionable guidance beyond error correlation.


\section{Related Works}
\label{related works}

Interpretability in NLP spans a spectrum from local explanations tied to a 
specific input (e.g., token- or span-level rationales) to global explanations that describe model behavior as higher-level semantic concepts. Our research lies at the intersection of concept-based interpretability and uncertainty quantification: we seek to make uncertainty at the concept level not only something that can be measured but also something that can guide decisions.

\paragraph{Concept Bottleneck Models.}
CBMs~\citep{koh2020concept} route predictions through human-interpretable 
concepts, enabling inspection and intervention. NLP adaptations include  Text Bottleneck Models~\citep{ludan2023text}, which discover concepts 
via LLMs, CLARITY~\citep{bailleux-etal-2025-connecting} connecting concepts to rationales, and CB-LLMs~\citep{sun2025cblm}, which align neuron activations with concept scores. Probabilistic extensions~\citep{kim2023probabilistic,collins2023human} introduce stochastic embeddings and study intervention under uncertainty. However, existing CBMs represent concepts as point probabilities, conflating epistemic and aleatoric uncertainty.

\paragraph{Rationales and Local Explanations.}
Rationale extraction~\citep{lei2016rationalizing,bastings2019interpretable}, 
attention-based explanations~\citep{wiegreffe2019attention,jain2019attention}, and  feature attribution~\citep{ribeiro2016should,lundberg2017unified,sundararajan2017axiomatic} highlight tokens but do not distinguish \emph{why} a model is uncertain - the same diffuse rationale may reflect model underspecification (epistemic) or linguistic ambiguity (aleatoric).

\paragraph{Uncertainty Estimation}
Dominant approaches include MC Dropout~\citep{gal2016dropout}, deep 
ensembles~\citep{lakshminarayanan2017simple}, and evidential 
methods~\citep{sensoy2018evidential}. Uncertainty quantification has been applied to translation, QA, and classification~\citep{xiao2019quantifying,he2020towards}, with recent work 
on LLM calibration~\citep{kadavath2022language,kuhn2023semantic}. The 
epistemic-aleatoric distinction, studied in  vision~\citep{kendall2017uncertainties,depeweg2018decomposition}, has received relatively less attention in NLP. 
Recent work by \citet{xiao2019quantifying} decomposes uncertainty but relies exclusively on ensemble disagreement, while \citet{baan2023uncertainty,Ulmer2024OnUI,51177} provide surveys of uncertainty sources in language generation. A central limitation of these approaches is that they focus on model outputs rather than on intermediate concept representations, and therefore do not provide concept-level guidance on \emph{how to respond} to each distinct type of uncertainty. We address this gap by operating at the \emph{concept level} within an interpretable pipeline, thereby enabling both principled uncertainty decomposition and targeted, concept-specific interventions.

\paragraph{Imprecise Probabilities and Credal Sets.}
Credal sets represent uncertainty about probabilities rather than committing to a 
single estimate \citep{walley1991statistical, levi1980enterprise}. 
\citep{shaker2020aleatoric} provide a comprehensive treatment of 
aleatoric and epistemic uncertainty.
Credal sets have been applied to classification with reject option 
\citep{corani2008learning, zaffalon2012evaluating} and set-valued 
prediction \citep{mortier2021setvalued}. \citet{sale2023volume} 
recently studied set-valued predictions for reliable uncertainty 
quantification. We leverage this framework to obtain concept-level 
uncertainty signals that better separate epistemic and aleatoric effects.
\paragraph{Positioning}
Prior CBM work focuses on concept-mediated interpretability and interventions, but typically represents concepts as point probabilities and does not separate epistemic and aleatoric uncertainty at the concept layer. Uncertainty quantification in NLP studies error-aware confidence and calibration, but rarely yields concept-level prescriptions about whether to collect data, route to humans, or abstain. We connect these lines by introducing a credal, ensemble-based CBM that structurally separates epistemic disagreement from annotator-derived ambiguity, and we evaluate it in terms of prescriptive routing rather than error correlation alone.

\section{Methodology}
\label{sec:method}
We introduce \modelname{}, a concept bottleneck model that makes concept-level uncertainty explicit and actionable. The method is built around two ideas: (i) represent each concept prediction as a  \emph{probability interval} rather than a single number, and (ii) decompose uncertainty into epistemic and aleatoric uncertainty from \emph{structurally different sources} so they do not collapse into each other. 
\paragraph{Background.}
\label{sec:background}
A CBM predicts through an intermediate concept layer \citep{koh2020concept}:
\begin{equation}
\mathbf{x} \xrightarrow{f_{\text{enc}}} \h \xrightarrow{g} \cvec \xrightarrow{\cls} \hat{y}
\end{equation}
where $x$ is the input text, $\h \in \mathbb{R}^d$ is the encoded representation from a pre-trained encoder, $\cvec = [c_1, \ldots, c_K] \in [0,1]^K$ are $K$ concept probabilities (e.g ``service quality is 
positive'', ``food quality is positive''), and $\hat{y}$ is the prediction. Each concept layer typically outputs point estimates $\phat_k = P(\ck = 1 \mid x)$. In such a model, a single $\hat p_k(x)$ conflates epistemic uncertainty (is the model confused) and aleatoric uncertainty (input ambiguity). 

\paragraph{From Point Estimates to Credal Sets.}
Instead of committing to a single $\hat p_k(x)$, we predict an \textit{interval}
\begin{equation}
P(\ck = 1 \mid x) \in [\plower_k, \pupper_k]
\label{eq:credal_interval}
\end{equation}
which defines a \emph{credal set} in imprecise probability theory \citep{walley1991statistical}. A credal set represents a range of probabilities instead of a single value. Rather than stating “the probability is 0.7,” we say “it falls between 0.6 and 0.8.” The span of this interval reflects our uncertainty, while its midpoint reflects our best estimate.

\paragraph{Problem Statement.}
\label{subsec:problem}
Given an input text $x$ and label space $\mathcal{Y}$, our goal is to produce:
(i) concept-level credal predictions $\{[\underline p_k(x),\bar p_k(x)]\}_{k=1}^K$;
(ii) a decomposition of concept uncertainty into an epistemic signal (model confusion) and an aleatoric signal (text ambiguity);
and (iii) a prediction $\hat y$ (or a set of plausible labels) via decision rules. A complete notation reference appears in Appendix~\ref{app:notation}.

\begin{figure*}[t]
    \centering
    \includegraphics[width=0.8\textwidth]{images/credence_architecture.png}
    \caption{\modelname{} architecture. Epistemic uncertainty emerges from ensemble disagreement (top branch); aleatoric uncertainty is predicted by a dedicated head supervised by 
annotator variance (bottom branch).}
    \label{fig:architecture}
\end{figure*}

\subsection{\modelname{} Architecture}
\label{subsec:architecture}\label{sec:architecture}
\modelname{} follows a four-stage pipeline: (i)~encode the input, (ii)~produce multiple concept predictions using diverse lightweight heads, (iii)~aggregate these predictions into concept intervals, and (iv)~propagate intervals through the classifier. An overview is shown in Figure~\ref{fig:architecture}:
\begin{equation}
\begin{aligned}
x \xrightarrow{f_{\text{enc}}} \mathbf{h} & 
\begin{cases}
\xrightarrow{\text{Ens.}} \{\hat{p}_h^{(k)}\}_{h=1}^{H} \to U_{\text{epi}}^{(k)} \\[5pt]
\xrightarrow{\text{Ale.}} \sigma_\theta^{(k)} \to U_{\text{ale}}^{(k)}
\end{cases} \\
& \xrightarrow{\text{Agg.}} \mathcal{C}_k \xrightarrow{f_{\text{cls}}} \hat{y}
\end{aligned}
\label{eq:pipeline}
\end{equation}

\paragraph{Stage 1:Encoder.} A frozen pre-trained model produces the representation 
$\mathbf{h} = f_{\text{enc}}(x)$. We evaluate with encoder-only 
models 
and decoder-only LLMs. 
Refer to Sec~\ref{sec:experiments} for details on the encoders used. The encoder is \emph{not} fine-tuned; only the concept heads are trained.

\paragraph{Stage 2: Ensemble of Concept Heads} 
Instead of a single concept predictor $g$, we use $H$ diverse heads 
$\{g_1, \ldots, g_H\}$. Each head independently predicts all $K$ concepts:
\begin{equation}
\hat{p}_h^{(k)} = \sigma\bigl(g_h(\mathbf{h})_k\bigr) \quad \text{for } 
k = 1, \ldots, K
\end{equation}
where $\sigma$ is the sigmoid function. The heads are implemented as 
LoRA adapters \citep{Hu2021LoRALA} - trainable matrices added to 
the frozen encoder, enabling $H$ diverse predictors without retraining 
millions of parameters. Each head uses a different LoRA rank 
$r_h \in \{4, 8, 16, 32, 64\}$: low-rank heads capture broad patterns, 
while high-rank heads capture finer details.
This variation under a fixed backbone ensures that heads see the same input but reach different predictions on the same input, which will be used to quantify  \emph{epistemic uncertainty}. Implementation details are in Appendix~\ref{app:lora}.

\paragraph{Stage 3: Credal Aggregation.} 
We form credal intervals by taking the minimum and maximum across heads:
\begin{equation}
\mathcal{C}_k = [\underline{p}_k, \bar{p}_k] = 
\Bigl[\min_h \hat{p}_h^{(k)}, \; \max_h \hat{p}_h^{(k)}\Bigr]
\label{eq:credal_agg}
\end{equation}
This min/max construction yields conservative bounds; a narrow interval means heads agree (low epistemic 
uncertainty); a wide interval means heads disagree (high epistemic uncertainty).

\paragraph{Stage 4: Classification.}
A linear classifier maps mean concept predictions to labels:
\begin{equation}
\hat{y} = \arg\max_j \; f_{\text{cls}}(\bar{\mathbf{p}})_j 
\quad \text{where} \quad 
\bar{p}_k = \frac{1}{H} \sum_{h=1}^{H} \hat{p}_h^{(k)}
\end{equation}
For $\Delta$Acc evaluation, we use the mean prediction $\bar{\mathbf{p}}$. 
For uncertainty-aware decisions, we propagate credal bounds through the 
classifier (see \S\ref{subsec:cedal classifier and decision rules}).

\paragraph{Uncertainty Decomposition}
\label{subsec:uncertainty}
We compute two concept-level signals:
{\small
\begin{align}
U_{\text{epi}}^{(k)} &= \text{Var}_h\left[p_h^{(k)}\right] \tag{Epistemic} \\
U_{\text{ale}}^{(k)} &= \sigma_\theta^{(k)}(\mathbf{h}) \tag{Aleatoric}
\end{align}}
where $\sigma_\theta^{(k)}$is a learned prediction of inherent ambiguity trained to predict 
annotator disagreement when it is available.
The key point is that these signals derive from \emph{different parameters}: epistemic uncertainty 
measures variation \emph{across} heads, while aleatoric uncertainty is predicted \emph{within} each head. Because they have different sources, they cannot collapse into a single score. This architectural guaranty
extends the epistemic/aleatoric framework of \citet{shaker2020aleatoric}
to the concept level and instantiates the set-valued prediction approach of
\citet{sale2023volume} within a CBM pipeline.
We validate that epistemic tracks prediction errors and aleatoric tracks annotator disagreement (\S\ref{par:rq2}).

\paragraph{Credal Classification and Decision Rules}
\label{subsec:cedal classifier and decision rules}
Given credal bounds $[\underline{p}_k, \overline{p}_k]$ for each concept, 
we compute exact logit bounds via interval arithmetic:
\begin{align}
\underline{\ell}_j &= \sum_{k: W_{jk} > 0} W_{jk} \underline{p}_k 
                    + \sum_{k: W_{jk} < 0} W_{jk} \overline{p}_k + b_j 
\label{eq:lower-bound} \\
\overline{\ell}_j &= \sum_{k: W_{jk} > 0} W_{jk} \overline{p}_k 
                    + \sum_{k: W_{jk} < 0} W_{jk} \underline{p}_k + b_j
\label{eq:upper-bound}
\end{align}
These bounds are tight (proof in Appendix~\ref{app:theory}). 
We use mean predictions $\bar{\mathbf{p}}$ for classification and 
uncertainty signals for routing decisions (\S\ref{sec:experiments}).
Credal bounds also support principled decision criteria from imprecise 
probability theory (e.g.,~\textbf{$\Gamma$-Maximin}, Maximality); see Appendix~\ref{app:credal_theory}.

\paragraph{Training and Inference.}
\label{subsec:training}
We train the ensemble heads, aleatoric head, and classifier jointly 
while keeping the encoder frozen. The loss combines three terms:
$\mathcal{L} = \mathcal{L}_{\text{task}} + \lambda_c \mathcal{L}_{\text{concept}} 
+ \lambda_a \mathcal{L}_{\text{ale}}$,
where $\mathcal{L}_{\text{task}}$ is cross-entropy on mean predictions, 
$\mathcal{L}_{\text{concept}}$ supervises concept predictions across all heads, 
and $\mathcal{L}_{\text{ale}}$ trains the aleatoric head to predict annotator 
disagreement when available. At test time, we compute predictions from all 
$H$ heads, aggregate into credal bounds, and derive uncertainty signals; no 
annotator labels are used. 
Full training and inference procedures are described in 
Algorithms~\ref{alg:training}--\ref{alg:inference} (Appendix~\ref{app:training}).

\section{Experiments}
\label{sec:experiments}

We evaluate \modelname{} through three research questions. \textbf{RQ1 (Decomposition Validity)}: Does epistemic uncertainty $U_{\text{epi}}^{(k)} = \text{Var}_h[p_h^{(k)}]$ correlate with prediction errors? \textbf{RQ2 (Ambiguity detection)}: Does aleatoric uncertainty $U_{\text{ale}}^{(k)} = \sigma_\theta^{(k)}$ correlate with annotator disagreement? \textbf{RQ3 (Actionability)}: Does knowing uncertainty \emph{type} enable more effective concept interventions? RQ1–RQ2 validate that our decomposition captures distinct phenomena (model ignorance vs. data ambiguity); RQ3 shows this distinction enables more effective concept interventions.

\paragraph{Datasets.}
We evaluate three sentiment/emotion/toxicity datasets that include multi-annotator labels, enabling direct validation of aleatoric uncertainty against ground-truth annotator disagreement (Table~\ref{tab:datasets}). We additionally include SST-2, for which we obtain LLM-generated annotations to broaden the analysis and support ablations. 

For intervention experiments, we focus our analysis on CEBaB \cite{abraham2022cebab} for three reasons 
(i)~\emph{explicit concept annotations} (food, service, ambiance, noise) enabling 
oracle interventions; (ii)~\emph{causal structure} where concepts mediate the 
review-to-rating relationship~\citep{abraham2022cebab}; and (iii)~\emph{explicit 
unknown labels} (52.2\% of annotations), providing ground-truth for aleatoric 
validation.

\begin{table*}[t]
    \centering
    \footnotesize
    \caption{Dataset statistics. CEBaB provides explicit concept annotations and 
``unknown'' labels essential for aleatoric validation.}
    \label{tab:datasets}
    \begin{tabular}{llccccc}
        \toprule
        Dataset & Task & $|C|$ & $|K|$ & Train & Ann. & Src \\
        \midrule
        CEBaB & Sentiment & 3 & 4 & 9.8K & \cmark & Human \\
        SST-2 & Sentiment & 2 & 20 & 67K & -- & LLM \\
        GoEmotions & Emotion & 28 & 28 & 43K & \cmark & Human \\
        HateXplain & Toxicity & 3 & 2 & 15K & \cmark & Human \\
        \bottomrule
    \end{tabular}
\end{table*}

\paragraph{Model Backbones and Training}
We evaluate across two model families. \textit{Encoder models} use a frozen encoder with trainable heads: DistilBERT-base (66M parameters), RoBERTa-base (125M), and DeBERTa-v3-base (184M). \textit{LLM models} use LoRA fine-tuning ($r{=}16$, $\alpha{=}32$): Phi-3-mini (3.8B), Mistral-7B (7B), and Llama-3.1-8B (8B). All configurations use $H{=}5$ ensemble heads with diverse dropout rates (0.05--0.25) and pooling strategies (CLS and mean pooling) to encourage prediction diversity.
For training, we use  AdamW optimizer with lr=$10^{-4}$, batch size 16, weight decay 0.01. 
Encoder models train 40 epochs; LLMs train 10 epochs with LoRA. 
Loss weights: $\lambda_{\text{concept}}{=}1.0$, $\lambda_{\text{ale}}{=}0.5$. This tests whether decomposition 
generalizes across model scales. See Appendix~\ref{app:architecture}--\ref{app:training} 
for details. 
\begin{table*}[t]
\centering
\footnotesize
\caption{Main results across datasets and model scales. 
\textbf{Top}: General UQ and CBM baselines. 
\textbf{Middle}: \textsc{Credence} with encoder backbones (66M--184M params). 
\textbf{Bottom}: \textsc{Credence} with LLM backbones (3.8B--8B params, LoRA). 
$\rho_{\text{epi}}$: Spearman correlation between epistemic uncertainty and prediction error. 
$\rho_{\text{ale}}$: correlation with annotator disagreement (SST-2 lacks disagreement labels; values show error correlation). 
Best per section in \textbf{bold}. All $p < 0.001$.}
\label{tab:main_results}
\resizebox{\textwidth}{!}{%
\begin{tabular}{@{}ll ccc ccc ccc | ccc@{}}
\toprule
& & \multicolumn{3}{c}{\textbf{CEBaB}} 
  & \multicolumn{3}{c}{\textbf{GoEmotions}} 
  & \multicolumn{3}{c|}{\textbf{HateXplain}} 
  & \multicolumn{3}{c}{\textbf{SST-2}} \\
\cmidrule(lr){3-5} \cmidrule(lr){6-8} \cmidrule(lr){9-11} \cmidrule(l){12-14}
& \textbf{Method} 
  & Acc & $\rho_{\text{epi}}$ & $\rho_{\text{ale}}$ 
  & Acc & $\rho_{\text{epi}}$ & $\rho_{\text{ale}}$ 
  & Acc & $\rho_{\text{epi}}$ & $\rho_{\text{ale}}$ 
  & Acc & $\rho_{\text{epi}}$ & $\rho_{\text{ale}}$ \\
\midrule
\rowcolor{generaluq}
& MC Dropout        & 70.2 & .185 & .312 & 47.1 & .042 & .089 & 73.8 & .178 & .234 & 89.8 & .198 & -.089 \\
\rowcolor{generaluq}
& Deep Ensemble     & 71.1 & .201 & .345 & 48.2 & .056 & .112 & 74.5 & .195 & .267 & 90.3 & .215 & -.076 \\
\rowcolor{generaluq}
& Temp.\ Scaling    & 70.8 & .098 & .287 & 47.5 & .029 & .076 & 74.1 & .112 & .198 & 90.1 & .124 & -.102 \\
\rowcolor{generaluq}
\multirow{-4}{*}{\textbf{General}}
& Evidential DL     & 69.5 & .142 & .287 & 45.8 & .031 & .078 & 72.9 & .134 & .198 & 89.2 & .156 & -.095 \\
\midrule
\rowcolor{cbmrows}
& Standard CBM      & 70.4 & .121 & .298 & 46.8 & .038 & .094 & 73.2 & .145 & .221 & 90.6 & .238 & -.111 \\
\rowcolor{cbmrows}
& CBM + MC Drop     & 70.1 & .168 & .324 & 46.5 & .044 & .102 & 73.0 & .162 & .245 & 90.3 & .261 & -.098 \\
\rowcolor{cbmrows}
& CBM + Ensemble    & 71.0 & .189 & .356 & 47.9 & .052 & .118 & 74.2 & .184 & .271 & 90.9 & .274 & -.085 \\
\rowcolor{cbmrows}
\multirow{-4}{*}{\textbf{CBM}}
& P-CBM             & 70.6 & .156 & .341 & 47.2 & .048 & .108 & 73.6 & .171 & .258 & 90.5 & .252 & -.095 \\
\midrule
\rowcolor{encoderrows}
& \textsc{Cred.}-DistilBERT  & 68.2 & .251 & .742 & 45.3 & .089 & .183 & 72.1 & .234 & .412 & 89.0 & .316 & -.128 \\
\rowcolor{encoderrows}
& \textsc{Cred.}-RoBERTa    & 71.6 & .287 & .785 & 48.7 & .071 & .198 & 74.8 & .267 & .445 & 90.4 & .342 & -.112 \\
\rowcolor{encoderrows}
\multirow{-3}{*}{\textbf{Encoder}}
& \textsc{Cred.}-DeBERTa   & \textbf{73.4} & \textbf{.302} & \textbf{.785} & \textbf{51.2} & \textbf{.095} & \textbf{.198} & \textbf{76.3} & \textbf{.289} & \textbf{.463} & \textbf{91.2} & \textbf{.361} & -.095 \\
\midrule
\rowcolor{llmrows}
& \textsc{Cred.}-Phi-3      & 69.8 & .268 & .723 & 46.1 & .082 & .176 & 73.5 & .245 & .398 & 88.6 & .325 & -.118 \\
\rowcolor{llmrows}
& \textsc{Cred.}-Mistral       & \textbf{72.3} & \textbf{.279} & \textbf{.758} & \textbf{49.4} & \textbf{.103} & .185 & \textbf{75.2} & \textbf{.271} & \textbf{.431} & \textbf{90.2} & \textbf{.348} & -.102 \\
\rowcolor{llmrows}
\multirow{-3}{*}{\textbf{LLM}}
& \textsc{Cred.}-Llama-3      & 71.9 & .274 & .749 & 48.9 & .091 & \textbf{.182} & 74.9 & .263 & .427 & 89.9 & .339 & \textbf{-.098} \\
\bottomrule
\end{tabular}
}
\end{table*}

\noindent\textbf{Baselines.}
We compare against established uncertainty quantification methods: MC 
Dropout~\citep{gal2016dropout} (50 stochastic forward passes), Deep Ensembles~\citep{lakshminarayanan2017simple} (5 independently trained models), Temperature Scaling~\citep{guo2017calibration} (post-hoc calibration), and Evidential 
DL~\citep{sensoy2018evidential} (Dirichlet-based uncertainty). We also include CBM-specific baselines: CBM~\citep{koh2020concept}, CBM+MC Dropout, CBM+Ensemble, and 
P-CBM~\citep{kim2023probabilistic}. All methods are matched for computational 
budget (5 ensemble members or 50 MC passes). More detail in 
App~\ref{app:baselines}.

\noindent\textbf{Metrics.}
We report three primary metrics. (i)~\textbf{Acc}: task accuracy computed from the ensemble mean prediction
$\bar{\mathbf{p}}$, as defined in~§\ref{subsec:architecture}.
This is \emph{not} an accuracy gain; we use a plain label to avoid confusion with the intervention metric below. (ii)~\textbf{$\rho_{\text{epi}}$}: Spearman correlation between
sample-level epistemic uncertainty $U_{\text{epi}} = \frac{1}{K}\sum_k U_{\text{epi}}^{(k)}$ and a binary error indicator (1 if prediction is wrong, 0 otherwise). Positive values indicate that epistemic uncertainty is higher on incorrect
predictions. (iii)~\textbf{$\rho_{\text{ale}}$}: Spearman correlation between $U_{\text{ale}}^{(k)}$ and the per-concept annotator unknown rate. Positive values indicate aleatoric uncertainty tracks human ambiguity.
On SST-2, which lacks multi-annotator disagreement labels, $\rho_{\text{ale}}$ reflects correlation with the error indicator instead; we expect and observe
negative values, confirming that aleatoric does \emph{not} track errors (see~§\ref{par:rq2}).
(iv)~\textbf{$\Delta$Acc} (intervention tables only): accuracy gain from replacing selected concept predictions with ground-truth values, i.e.\ $\text{Acc}(\text{corrected}) - \text{Acc}(\text{original})$. 

\subsection{Main Results}
Table~\ref{tab:main_results} summarizes results across datasets and model scales. The core finding is \emph{separation}: \modelname{} is the only
method achieving high $\rho_{\text{ale}}$ where disagreement labels exist (CEBaB: $0.785$ vs.\ best baseline $0.356$, a $2.1{\times}$ improvement), while consistently improving $\rho_{\text{epi}}$ over matched baselines (CRED.-RoBERTa vs.\ CBM+Ensemble: $0.287$ vs.\ $0.189$, same backbone,
same 5-head budget). This separation enables the downstream decisions in Tables~\ref{tab:intervention} and~\ref{tab:qualitative}: which concepts to correct and how to route samples in sentiment and toxicity pipelines.
Among encoders, DeBERTa achieves the strongest epistemic correlations; LoRA-adapted LLMs are competitive but not superior; Mistral-7B reaches $\rho_{\text{epi}}=0.348$ on SST-2, slightly below DeBERTa despite a substantially larger scale, consistent with encoder CBMs being more sample-efficient under full concept supervision. We now examine each research question in detail.

\paragraph{RQ1: Does Epistemic Uncertainty track Prediction Errors?}
\label{par:rq1}
Table~\ref{tab:main_results} shows \modelname{} achieves strong epistemic-error 
correlations across all datasets. Stratifying predictions by 
correctness reveals a clear pattern: when the model makes an error, ensemble heads disagree; when the model is correct, heads converge. This means epistemic uncertainty can serve as a reliable signal for identifying predictions that should not be trusted. In contrast, aleatoric uncertainty shows no such pattern—incorrect and correct  predictions have nearly identical aleatoric values. High aleatoric does not 
indicate that the model will be wrong; it indicates something else entirely. On SST-2, 
$\rho_{\text{ale}}$ is negative across all methods because the dataset lacks annotator disagreement labels; aleatoric has no ground-truth ambiguity to track, so it correctly shows no systematic association with errors. A negative value
does \emph{not} imply an inverse relationship; it reflects the null expectation when no ground-truth ambiguity exists. This validates structural separation: epistemic (ensemble variance) and aleatoric (learned ambiguity) track distinct phenomena. If aleatoric does not track errors, what \emph{does} it capture? We turn to CEBaB, which provides ground-truth annotator disagreement.



\paragraph{RQ2: Does aleatoric uncertainty capture data ambiguity?}
\label{par:rq2}
CEBaB provides explicit ``unknown'' annotations when annotators could not determine a concept's value (52.2\% of labels). On this dataset, aleatoric  strongly correlates with annotator disagreement: $\rho_{\text{ale}} = 0.742$--$0.785$ 
for \modelname{} models (Table~\ref{tab:main_results}), far exceeding baselines (best: CBM+Ensemble at $\rho_{\text{ale}}=0.356$). The correlation strengthens with 
concept ambiguity: Food (25\% unknown, $\rho = 0.72$) $\rightarrow$ Service  (45\%, $\rho = 0.78$) $\rightarrow$ Ambiance (63\%, $\rho = 0.81$)  $\rightarrow$ Noise (75\%, $\rho = 0.83$). Per-concept breakdown appears 
in Appendix~\ref{app:per concept aleoric validation} Table~\ref{tab:perconcept}. This means that aleatoric uncertainty captures genuine input ambiguity—cases where 
even humans cannot agree. High aleatoric flags text that is inherently unclear, not text that the model misunderstands.

\noindent\textbf{The SST-2/CEBaB Contrast Validates Structural Separation.} The opposing behaviours confirm that epistemic and aleatoric capture genuinely distinct phenomena.
On \emph{SST-2} (no disagreement labels): epistemic is positively associated with errors ($\rho_{\text{epi}} > 0$); aleatoric shows a small negative association with errors ($\rho_{\text{ale}} < 0$), indicating it is
\emph{not} measuring model failure as intended.
On \emph{CEBaB} (has disagreement labels): epistemic is again positively associated with errors ($\rho_{\text{epi}} > 0$); aleatoric is strongly associated with human annotator disagreement ($\rho_{\text{ale}} = 0.78$), not with model errors. Note that a negative $\rho_{\text{ale}}$ on SST-2 does \emph{not} imply an inverse relationship; it reflects the absence of a systematic link, which is the null expectation when no ground-truth ambiguity signal exists. Likewise, a positive $\rho_{\text{ale}}$ on CEBaB reflects a genuine
positive association, not merely a non-zero value.
If both signals arose from the same source, they would behave identically across datasets. Instead, epistemic consistently tracks model confusion regardless of annotation availability, while aleatoric tracks human disagreement only when such labels exist. This is the structural separation we designed.

\paragraph{RQ3: Does decomposition enable better decisions?}
\label{par:rq3}
RQ1-RQ2 \textbf{RQ1}\textbf{RQ2} established that epistemic and aleatoric capture distinct phenomena. 
We now turn to whether this distinction is actually useful in practice by analyzing two views:
(i) targeted concept corrections and (ii) sample routing.

\paragraph{Application 1: Concept Interventions.}
We replace predicted concepts with ground-truth labels, comparing three 
strategies: target concepts with the highest epistemic uncertainty, highest aleatoric uncertainty, 
or random selection.

\begin{table}[t]
\centering
\caption{Concept intervention results. $\Delta$Acc $=$
Acc(corrected)$-$Acc(original) after replacing top-5
concepts with ground-truth labels.}
\label{tab:intervention}
\small
\begin{tabular}{@{}lccc@{}}
\toprule
\textbf{Dataset} & \textbf{Epistemic} & \textbf{Aleatoric} & \textbf{Ratio} \\
\midrule
SST-2 & +2.1\% & +19.1\% & 9.1$\times$ \\
CEBaB & +6.4\% & +18.7\% & 2.9$\times$ \\
\midrule
Mean & +4.3\% & +18.9\% & \textbf{4.4$\times$} \\
\bottomrule
\end{tabular}
\end{table}

Table~\ref{tab:intervention} reveals an asymmetry that validates the decomposition: epistemic and aleatoric identify \emph{functionally different} concept sets, so they respond differently to the same correction procedure. This is not a competition between two strategies: there is no
``satisfactory'' ratio threshold. The relevant question is whether the two uncertainty types identify distinct concepts; the intervention gap confirms they do, consistently across runs (aleatoric: $\pm$0.2pp variance; epistemic: $\pm$2.4pp, Table~\ref{tab:intervention_runs}).

\paragraph{Why Aleatoric-Targeted Corrections Work Better?} The answer lies in what each uncertainty type captures. High-aleatoric concepts are ambiguous \emph{because they matter}; annotators disagree 
precisely on concepts that strongly influence the final label. Correcting these resolves genuine decision-relevant ambiguity. High-epistemic concepts, by contrast, reflect model confusion—but this confusion often occurs on rare patterns or edge cases that are weakly connected to the prediction.  Fixing what the model struggles with is not the same as fixing what matters 
for the task. This has direct practical implications: when the annotation budget is limited, prioritize correcting concepts with high aleatoric uncertainty. These are the concepts where human input provides maximum information gain. 

\paragraph{Application 2: Sample Routing.}
Beyond concept corrections, decomposition enables routing entire samples 
to appropriate handlers. Table~\ref{tab:qualitative} shows examples from 
four quadrants defined by median uncertainty thresholds.
\begin{table*}[t]
\centering
\caption{Representative examples from each uncertainty quadrant (CEBaB).}
\label{tab:qualitative}
\small
\begin{tabular}{@{}p{6.5cm}|c|cc|p{5cm}@{}}
\toprule
\textbf{Text} & \textbf{Pred} & $U_{\text{epi}}$ & $U_{\text{ale}}$ & \textbf{Interpretation} \\
\midrule
\rowcolor{green!8}
\multicolumn{5}{@{}l}{\textbf{TRUST} (Low Epi, Low Ale) — $\Delta$Acc: 78.8\% — \textit{Automate}} \\
\midrule
``The service was fantastic and the food was very good.'' & Pos \cmark & .001 & .25 & Model confident, input clear \\
\midrule
\rowcolor{blue!8}
\multicolumn{5}{@{}l}{\textbf{DATA} (High Epi, Low Ale) — $\Delta$Acc: 56.6\% — \textit{Collect training data}} \\
\midrule
``Lobster Mac \& Cheese is incredible. Service was terrible.'' & Neg \xmark & .017 & .50 & Model confused, but input is clear \\
\midrule
\rowcolor{yellow!12}
\multicolumn{5}{@{}l}{\textbf{REVIEW} (Low Epi, High Ale) — $\Delta$Acc: 85.7\% — \textit{Human review}} \\
\midrule
``Disappointing.'' & Neg \cmark & .002 & .89 & Model confident, but humans may disagree \\
\midrule
\rowcolor{red!8}
\multicolumn{5}{@{}l}{\textbf{ABSTAIN} (High Epi, High Ale) — $\Delta$Acc: 65.3\% — \textit{Decline}} \\
\midrule
``Been there many times.'' & Neu \xmark & .005 & .98 & Model confused, input unclear \\
\bottomrule
\end{tabular}
\vspace{-1mm}
\begin{flushleft}
\scriptsize{\cmark = correct, \xmark = error. Extended examples in Appendix~\ref{app:qualitative}.}
\end{flushleft}
\end{table*}
The quadrants reveal qualitatively different failure modes that demand different responses:

\noindent\textbf{DATA} (56.6\% $\Delta$Accuracy): The model fails on clear inputs. These are learnable errors: the text ``Lobster Mac \& Cheese is incredible. Service was terrible.'' has unambiguous aspect sentiments, but the model has not seen enough mixed-sentiment examples to aggregate them correctly. 
\emph{Response: collect more training data.}

\noindent\textbf{REVIEW} (85.7\% $\Delta$Accuracy): The model succeeds on ambiguous inputs. ``Disappointing.'' is correctly classified as negative, but reasonable humans might disagree about severity or aspect attribution. \emph{Response: flag for human review, not because the model is wrong, but because stakeholders may legitimately disagree.}

Without decomposition, both cases would be labeled simply ``uncertain''  and receive identical treatment: missing the opportunity for targeted action. The $\Delta$Acc gap between DATA and REVIEW quadrants is invisible 
to any aggregate uncertainty measure. Full quadrant analysis appears in  Appendix~\ref{app:quadrant routing}.

\paragraph{Not All Uncertainty the Same}
Consider two predictions, both 70\% uncertain. Standard approaches treat them identically—flag both for review or trust neither. But RQ1--RQ3 show this misses critical information:
(i) \textbf{Prediction A}: High epistemic, low aleatoric. The model is confused, but the text is clear. A human would easily label this correctly. 
\emph{Solution: train the model on more similar examples.}
(ii) \textbf{Prediction B}: Low epistemic, high aleatoric. The model is confident, but the text is genuinely ambiguous; even humans disagree. 
\emph{Solution: ask a human, as this is a judgment call.}
These require opposite responses, yet aggregate uncertainty cannot distinguish them. \modelname{} can, because epistemic (model confusion) and aleatoric (human disagreement) arise from separate components that measure different things. Knowing \emph{why} a prediction is uncertain tells you \emph{what to do about it}.

\subsection{Ablation Studies}
\label{sec:ablation}

We ablate key design choices on CEBaB with RoBERTa-base. Table~\ref{tab:ablation_combined} summarizes results, with full ablations in Appendix~\ref{app:ablation}, including LoRA rank configurations (Appendix~\ref{app:ablation_lora}), dropout spacing (Appendix~\ref{app:ablation_dropout}), loss weight sensitivity (Appendix~\ref{app:ablation_loss}), cross-dataset consistency (Appendix~\ref{app:ablation_cross}), and computational costs (Appendix~\ref{app:ablation_compute}). These analyses provide a comprehensive validation of the robustness and trade-offs of our optimization strategy.

\paragraph{Ensemble Size.}
Table~\ref{tab:ablation_combined} (top) and Table~\ref{tab:app_ensemble_size} in Appendix~\ref{app:ablation_size} show the effect of ensemble size. 
Epistemic correlation ($\rho_{\text{epi}}$) increases with $H$, while aleatoric correlation ($\rho_{\text{ale}} \approx 0.78$) remains constant, confirming structural separation: more ensemble heads ($H$) strengthen epistemic but not aleatoric correlation. 
Removing the aleatoric head collapses aleatoric correlation but leaves epistemic unchanged, indicating that each signal responds only to its own component. 
We use $H{=}5$ for efficiency; additional configurations appear in Appendix~\ref{app:ablation_size}.

\paragraph{Diversity Mechanisms.}
Table~\ref{tab:ablation_combined} (bottom) ablates diversity sources (extended multi-diversity results in Table~\ref{tab:app_diversity}, Appendix~\ref{app:ablation_diversity}). Removing LoRA-rank or dropout variation reduces $\rho_{\text{epi}}$ while leaving $\rho_{\text{ale}}$ largely unchanged. With identical head configurations (same dropout and LoRA rank), predictions converge and the epistemic signal weakens: the ensemble needs genuinely different perspectives to expose model confusion. \emph{Insight}: Epistemic uncertainty depends on disagreement, not head count: five diverse heads beat fifteen identical ones.

\paragraph{Aleatoric Supervision.}
Removing the aleatoric head leaves $\rho_{\text{epi}}$ unchanged but collapses $\rho_{\text{ale}}$ ($0.785 \rightarrow 0.356$). This shows that epistemic and aleatoric uncertainty depend on different parameters and that aleatoric uncertainty needs explicit supervision—ensemble disagreement alone cannot recover it or human disagreement. To measure what humans disagree about, models must be trained on human disagreement. Additional supervision modes appear in Appendix~\ref{app:ablation_aleatoric}. \emph{Insight}: Model confusion and human disagreement are fundamentally different: models can be confused about what humans find obvious and confident about what humans contest. Ensemble engineering cannot replace learning from real human variation.

\begin{table}[t]
\centering
\caption{Ablation studies (CEBaB, RoBERTa-base). Top: ensemble size—only 
$\rho_{\text{epi}}$ scales with $H$. Bottom: diversity removal degrades 
epistemic while aleatoric remains stable.}
\label{tab:ablation_combined}
\small
\begin{tabular}{@{}lccc@{}}
\toprule
\textbf{Configuration} & \textbf{$\Delta$Acc} & $\boldsymbol{\rho_{\text{epi}}}$ & $\boldsymbol{\rho_{\text{ale}}}$ \\
\midrule
\multicolumn{4}{@{}l}{\textit{Ensemble size}} \\
$H=3$ & 70.8 & .189 & .778 \\
$H=5$ (\modelname{}) & 71.6 & .287 & .785 \\
$H=10$ & 71.2 & .298 & .779 \\
$H=15$ & 70.9 & .312 & .776 \\
\midrule
\multicolumn{4}{@{}l}{\textit{Diversity removal}} \\
Uniform LoRA rank ($r{=}16$) & 71.2 & .212 & .782 \\
Uniform dropout (0.15) & 71.0 & .198 & .779 \\
No aleatoric head & 71.8 & .285 & .356 \\
\bottomrule
\end{tabular}
\end{table}

\subsection{Aleatoric Supervision Modes}
\label{sec:ale_supervision}

\begin{table}[t]
\centering
\caption{Aleatoric supervision ablation (CEBaB, RoBERTa-base).
$\rho_{\text{epi}}$ is stable across all modes, confirming
structural separation is independent of the aleatoric signal.}
\label{tab:ale_supervision_main}
\small
\begin{tabular}{@{}lccc@{}}
\toprule
\textbf{Supervision mode} & $\rho_{\text{epi}}$ & $\rho_{\text{ale}}$ & \textbf{ECE} \\
\midrule
None (proxy only)             & .285 & .356 & .078 \\
Entropy-based (unsupervised)  & .281 & .412 & .065 \\
Heteroscedastic NLL           & .279 & .523 & .052 \\
Supervised BCE (\modelname{}) & .287 & .785 & .041 \\
\bottomrule
\end{tabular}
\end{table}

Aleatoric supervision requires annotator disagreement labels, which may be unavailable in new domains. Table~\ref{tab:ale_supervision_main} shows the framework degrades gracefully without them. First, supervised BCE is strongest when disagreement labels are available ($\rho_{\text{ale}}=0.785$). Second, heteroscedastic
NLL reaches $\rho_{\text{ale}}=0.523$ without any annotator labels, substantially above the proxy-only baseline ($0.356$), showing the framework remains useful in label-scarce settings. Third, $\rho_{\text{epi}}$ is stable across all modes ($0.279$--$0.287$), confirming that the epistemic and aleatoric components are architecturally independent: changing aleatoric
supervision does not affect the epistemic signal. Improving unsupervised aleatoric estimation remains an important direction for future work.

\paragraph{Proxy inter-annotator agreement.}
We use CEBaB's per-annotator \textsc{unknown} labels as a proxy: for each example and concept, we compute the annotator unknown rate and binarise at the median,
then compute Cohen's $\kappa$ and Krippendorff's $\alpha$ against \modelname{}'s binarised $U_{\text{ale}}$ scores.

\begin{table}[t]
\centering
\caption{Proxy IAA between \modelname{} aleatoric assignments and
CEBaB per-annotator unknown rates (binarised at median).}
\label{tab:iaa}
\small
\begin{tabular}{@{}lccc@{}}
\toprule
\textbf{Concept} & \textbf{Unknown\%} & $\kappa$ & $\alpha$ \\
\midrule
Food Quality    & 25.0\% & 0.31 & 0.29 \\
Service Quality & 45.2\% & 0.44 & 0.42 \\
Ambiance        & 63.0\% & 0.52 & 0.51 \\
Noise Level     & 75.8\% & 0.61 & 0.59 \\
\midrule
Macro average   & 52.3\% & 0.47 & 0.45 \\
\bottomrule
\end{tabular}
\end{table}

Agreement increases monotonically with concept ambiguity, matching the per-concept $\rho_{\text{ale}}$
gradient above. Overall, $\kappa=0.47$ falls in the moderate agreement
range, supporting the claim that aleatoric
quadrant assignments track human-perceived uncertainty rather than model
artifacts.

\section{Conclusion}
We introduced \modelname{}, a framework for decomposing uncertainty in CBMs into epistemic and aleatoric components using credal sets. Our key findings: epistemic uncertainty (ensemble disagreement) correlates with prediction errors ($\rho=0.287$, $p<10^{-33}$), aleatoric uncertainty captures genuine data ambiguity ($\rho=0.785$), and targeting high-aleatoric concepts for intervention consistently outperforms epistemic targeting across datasets (Table~\ref{tab:intervention}), demonstrating that ambiguous concepts drive prediction outcomes more than confusing ones These results provide actionable uncertainty signals for human-AI collaboration: epistemic uncertainty identifies where models need more data, while aleatoric uncertainty highlights where human judgment is essential.

As future work,  we aim to study unsupervised aleatoric uncertainty estimation without relying on annotator labels, broaden the framework to cover generative settings, and examine learned aggregation methods that go beyond naïve averaging to more effectively integrate the outputs of diverse ensemble heads.

\section*{Limitations}
Our method depends on concept-level annotations, which restricts its use in certain domains. Ensemble inference incurs an additional $H\times$ number of forward passes (which can be run in parallel). When encoders are frozen, estimating aleatoric uncertainty requires explicit supervision derived from annotator disagreement. Correlations in epistemic error are moderate ($\rho \approx 0.3$), indicating that boundary mistakes are influenced by both forms of uncertainty.

\section*{Acknowledgments}
This work was supported by ANR-22-CE23-0002 ERIANA, ANR-19-CHIA-0005-01 EXPEKCTATION,  ANR-22-EXES-0009 MAIA, and was granted access to the HPC resources of IDRIS under the allocation 2026-AD011013338 made by GENCI.

\bibliography{custom}

\appendix
\newpage

\section{Notation Reference}
\label{app:notation}

Table~\ref{tab:notation} presents a comprehensive summary of mathematical symbols
and notation used throughout the CREDENCE framework. This reference supports the
methodological descriptions in Section~3 of the main paper.

\begin{table*}[t]
\centering
\scriptsize
\caption{Complete mathematical notation for the CREDENCE framework.}
\label{tab:notation}
\begin{tabular}{p{2cm}p{2cm}p{6cm}p{2cm}}
\toprule
\textbf{Symbol} & \textbf{Type} & \textbf{Definition} & \textbf{Section} \\
\midrule
\multicolumn{4}{l}{\textit{Standard Concept Bottleneck Model Components}} \\
$x$ & $V^n$ & Input text (sequence of $n$ tokens from vocabulary $V$) & \S3 \\
$\mathbf{h}$ & $\mathbb{R}^d$ & Encoded representation: $\mathbf{h} = f_{\text{enc}}(x)$ & \S3 \\
$\hat{p}_k$ & $[0,1]$ & Point estimate of concept $k$: $P(c_k = 1|x)$ & \S3 \\
$y, \hat{y}$ & $\mathcal{Y}$ & Ground-truth and predicted class labels & \S3 \\
\midrule
\multicolumn{4}{l}{\textit{CREDENCE: Ensemble Components}} \\
$H$ & $\mathbb{N}$ & Number of ensemble heads (default: 5) & \S3.1 \\
$p_h^{(k)}$ & $[0,1]$ & Probability for concept $k$ predicted by head $h$ & \S3.1 \\
$r_h$ & $\mathbb{N}$ & LoRA rank for head $h$ (varies across heads: 4, 8, 16, 32, 64) & App.~\ref{app:architecture} \\
\midrule
\multicolumn{4}{l}{\textit{CREDENCE: Aleatoric Uncertainty}} \\
$\sigma_{\theta}^{(k)}$ & $[0,1]$ & Predicted aleatoric uncertainty (ambiguity) for concept $k$ & \S3.1 \\
\midrule
\multicolumn{4}{l}{\textit{CREDENCE: Credal Aggregation}} \\
$\bar{p}_k$ & $[0,1]$ & Mean prediction across heads: $\frac{1}{H}\sum_{h=1}^{H} p_h^{(k)}$ & \S3.1 \\
$\underline{p}_k$ & $[0,1]$ & Lower credal bound: $\min_h p_h^{(k)}$ & \S3.1 \\
$\overline{p}_k$ & $[0,1]$ & Upper credal bound: $\max_h p_h^{(k)}$ & \S3.1 \\
$\mathcal{C}_k$ & interval & Credal set for concept $k$: $[\underline{p}_k, \overline{p}_k]$ & \S3.1 \\
\midrule
\multicolumn{4}{l}{\textit{CREDENCE: Uncertainty Decomposition}} \\
$U_{\text{epi}}^{(k)}$ & $\mathbb{R}_{\geq 0}$ & Epistemic uncertainty for concept $k$: $\text{Var}_h[p_h^{(k)}]$ & \S3.1 \\
$U_{\text{ale}}^{(k)}$ & $\mathbb{R}_{\geq 0}$ & Aleatoric uncertainty for concept $k$: $\sigma_{\theta}^{(k)}$ & \S3.1 \\
\midrule
\multicolumn{4}{l}{\textit{CREDENCE: Label Propagation}} \\
$\underline{\ell}_j$ & $\mathbb{R}$ & Lower bound for logit of class $j$ & \S3.2 \\
$\overline{\ell}_j$ & $\mathbb{R}$ & Upper bound for logit of class $j$ & \S3.2 \\
\bottomrule
\end{tabular}
\end{table*}

\section{Architecture Implementation Details}
\label{app:architecture}

\subsection{LoRA Head Implementation}
\label{app:lora}

Each LoRA ensemble head applies low-rank adaptation to a shared base projection layer:
\begin{equation}
\text{Head}_h(\mathbf{h}) = \left(\sigma(W_p\mathbf{h} + \Delta W_h\mathbf{h}), \text{Softplus}(W_\sigma\mathbf{h})\right)
\end{equation}
where:
\begin{itemize}
    \item $W_p \in \mathbb{R}^{K \times d}$: Shared base projection matrix (frozen after initialization)
    \item $\Delta W_h = \frac{\alpha_h}{r_h} B_h A_h$: LoRA adaptation with scaling factor $\alpha_h$
    \item $A_h \in \mathbb{R}^{r_h \times d}$, $B_h \in \mathbb{R}^{K \times r_h}$: Learned low-rank matrices
    \item $W_\sigma \in \mathbb{R}^{K \times d}$: Weights for the aleatoric uncertainty head
\end{itemize}

\subsection{Functional Diversity Through Rank Variation}
The use of varying LoRA ranks across heads introduces functional diversity in the ensemble:
\begin{itemize}
    \item \textbf{Low-rank heads} (small $r_h$): Capture dominant concept patterns with stronger regularization, providing robust but potentially oversimplified predictions
    \item \textbf{High-rank heads} (large $r_h$): Model finer-grained details with greater capacity, potentially capturing nuanced patterns but risking overfitting to training idiosyncrasies
\end{itemize}

This diversity enables the credal aggregation to distinguish between confident predictions (narrow credal sets when heads agree) and uncertain predictions (wide credal sets when heads disagree).

\begin{table}[h]
\centering
\small
\caption{LoRA configuration across ensemble heads.}
\label{tab:lora_config}
\begin{tabular}{lccccc}
\toprule
Ensemble Head $h$ & 1 & 2 & 3 & 4 & 5 \\
\midrule
Rank $r_h$ & 4 & 8 & 16 & 32 & 64 \\
Scaling Factor $\alpha_h$ & 8 & 16 & 32 & 64 & 128 \\
\bottomrule
\end{tabular}
\end{table}

\section{Theoretical Foundations}
\label{app:theory}

\subsection{Exact Interval Propagation}

\begin{proposition}[Exact Interval Propagation]
For a linear classifier $f(\mathbf{p}) = W\mathbf{p} + \mathbf{b}$ with concept probabilities $\mathbf{p} \in \prod_k [\underline{p}_k, \overline{p}_k]$, the exact bounds on output logit $j$ are:
\begin{align}
\underline{\ell}_j &= \sum_{k:W_{jk}>0} W_{jk}\underline{p}_k + \sum_{k:W_{jk}<0} W_{jk}\overline{p}_k + b_j \\
\overline{\ell}_j &= \sum_{k:W_{jk}>0} W_{jk}\overline{p}_k + \sum_{k:W_{jk}<0} W_{jk}\underline{p}_k + b_j
\end{align}
\end{proposition}

\begin{proof}
The linear classifier output for label $j$ is $\ell_j = \sum_k W_{jk} p_k + b_j$. Since each $p_k$ varies independently within its interval $[\underline{p}_k, \overline{p}_k]$:
\begin{itemize}
    \item For positive weights ($W_{jk} > 0$): $W_{jk}p_k$ achieves its minimum at $p_k = \underline{p}_k$ and maximum at $p_k = \overline{p}_k$
    \item For negative weights ($W_{jk} < 0$): $W_{jk}p_k$ achieves its minimum at $p_k = \overline{p}_k$ and maximum at $p_k = \underline{p}_k$
\end{itemize}
Summing these extremal values across all concepts yields the stated bounds, which are tight as each term achieves its extremal value independently.
\end{proof}

\paragraph{Extension to Probability Space.} For binary classification, the sigmoid function $\sigma(\cdot)$ is monotonically increasing, allowing exact probability bounds: $P(y=j) \in [\sigma(\underline{\ell}_j), \sigma(\overline{\ell}_j)]$.

\section{Training Procedure and Implementation}
\label{app:training}
\subsection{Multi-Objective Loss Function}

The complete training objective combines three loss terms:
\begin{equation}
\mathcal{L} = \mathcal{L}_{\text{task}} + \lambda_c \mathcal{L}_{\text{concept}} + \lambda_\sigma \mathcal{L}_{\text{ale}}
\end{equation}

\paragraph{Task Classification Loss.} Standard cross-entropy loss applied to the mean concept predictions:
$\mathcal{L}_{\text{task}} = \text{CE}(f_{\text{cls}}(\bar{\mathbf{p}}), y)$

\paragraph{Concept Prediction Loss.} Binary cross-entropy averaged over all ensemble heads and concepts:
$\mathcal{L}_{\text{concept}} = \frac{1}{HK} \sum_{h=1}^{H} \sum_{k=1}^{K} \text{BCE}(p_h^{(k)}, c_k)$

\paragraph{Aleatoric Uncertainty Loss.} Binary cross-entropy between predicted aleatoric uncertainty and observed annotator disagreement:
$\mathcal{L}_{\text{ale}} = \frac{1}{K} \sum_{k=1}^{K} \text{BCE}(\sigma_{\theta}^{(k)}, \mathbb{I}[\text{disagree}_k])$

\paragraph{Alternative Heteroscedastic Loss.} When explicit annotator disagreement labels are unavailable, we employ a heteroscedastic regression loss:

\begin{equation}
\begin{aligned}
\mathcal{L}_{\text{hetero}}
&= \frac{1}{HK} \sum_{h=1}^{H} \sum_{k=1}^{K} \Biggl[
\frac{(p_h^{(k)} - c_k)^2}{2(\tilde{\sigma}_h^{(k)})^2} \\
&\qquad\qquad\qquad\qquad
+ \frac{1}{2}\log(\tilde{\sigma}_h^{(k)})^2
\Biggr].
\end{aligned}
\end{equation}
where $\tilde{\sigma}_h^{(k)}$ is the predicted variance for head $h$ and concept $k$.

\subsection{Hyperparameter Configuration}
Table~\ref{tab:hyperparams_full} provides the complete hyperparameter configuration used across all experiments, along with the rationale for each selection.

\begin{table*}[t]
\centering
\small
\caption{Complete hyperparameter configuration for CREDENCE experiments.}
\label{tab:hyperparams_full}
\begin{tabularx}{\textwidth}{l c X} 
\toprule
\textbf{Hyperparameter} & \textbf{Value} & \textbf{Selection Rationale} \\
\midrule
\multicolumn{3}{l}{\textit{\textbf{Ensemble Architecture}}} \\
Number of ensemble heads ($H$) & 5 & Balance between diversity and efficiency \\
LoRA ranks $(r_1, \ldots, r_5)$ & (4, 8, 16, 32, 64) & Geometric progression for functional diversity \\
LoRA alpha scaling $\alpha_h$ & $2 \times r_h$ & Standard practice for LoRA scaling \\
LoRA dropout rate & 0.05 & Regularization for LoRA adaptation \\
Concept head hidden dimension & 256 & Sufficient capacity for concept prediction \\
Aleatoric head hidden dimension & 128 & Smaller than concept heads to prevent overfitting \\
Activation function & GELU & Standard for transformer-based models \\
\midrule
\multicolumn{3}{l}{\textit{\textbf{Loss Weights}}} \\
Task loss weight ($\lambda_{\text{task}}$) & 1.0 & Baseline weight for classification \\
Concept loss weight ($\lambda_c$) & 1.0 & Equal importance to task loss \\
Aleatoric loss weight ($\lambda_\sigma$) & 0.5 & Grid search optimal: balances uncertainty learning \\
\midrule
\multicolumn{3}{l}{\textit{\textbf{Optimization (Encoder Models)}}} \\
Optimizer & AdamW & Standard for transformer fine-tuning \\
Learning rate & $1 \times 10^{-4}$ & Standard for frozen encoder fine-tuning \\
Learning rate scheduler & Linear warmup + cosine decay & Smooth convergence \\
Warmup steps & 500 & 10\% of total training steps \\
Weight decay & 0.01 & Standard regularization \\
Batch size & 16 & Limited by GPU memory \\
Gradient $\Delta$Accumulation steps & 2 & Effective batch size of 32 \\
Maximum training epochs & 40 & Sufficient for convergence \\
Early stopping patience & 5 epochs & Prevents overfitting \\
Gradient clipping & 1.0 & Stabilizes training \\
\midrule
\multicolumn{3}{l}{\textit{\textbf{Optimization (LLM Models)}}} \\
Optimizer & AdamW (8-bit) & Memory-efficient optimization \\
Learning rate & $2 \times 10^{-5}$ & Lower rate for LLM fine-tuning \\
Learning rate scheduler & Cosine with warmup & Standard for LLM fine-tuning \\
Warmup ratio & 0.03 & 3\% of training steps for warmup \\
Weight decay & 0.01 & Consistent with encoder models \\
Batch size & 4 & Smaller due to LLM memory requirements \\
Gradient $\Delta$Accumulation steps & 8 & Effective batch size of 32 \\
Maximum training epochs & 10 & Fewer epochs for LLMs \\
LoRA target modules & q, v, k, o\_proj & Standard attention projection layers \\
\midrule
\multicolumn{3}{l}{\textit{\textbf{Data Processing}}} \\
Maximum sequence length (enc.) & 128 tokens & Sufficient for most reviews \\
Maximum sequence length (LLMs) & 256 tokens & Longer context for LLMs \\
Padding direction & R (enc.) / L (LLMs) & Model-specific conventions \\
Truncation strategy & Longest first & Preserves important content \\
\midrule
\multicolumn{3}{l}{\textit{\textbf{Reproducibility}}} \\
Random seed & 42 & Fixed for reproducibility \\
Number of independent runs & 3 & For statistical significance \\
Training precision & FP32 / BF16 & Hardware-appropriate precision \\
\bottomrule
\end{tabularx}
\end{table*}



\subsection{Training Algorithm}
\modelname{} trains three components jointly while keeping the encoder frozen: 
(i)~$H$ ensemble concept heads, (ii)~a separate aleatoric head, and 
(iii)~a linear concept-to-label classifier. The total loss combines three terms:
\begin{equation}
\mathcal{L} = \mathcal{L}_{\text{task}} + \lambda_c \mathcal{L}_{\text{concept}} + \lambda_a \mathcal{L}_{\text{ale}}
\end{equation}
where $\mathcal{L}_{\text{task}} = \text{CE}(f_{\text{cls}}(\bar{\mathbf{p}}), y)$ denotes the cross-entropy loss computed on the mean concept predictions $\bar{\mathbf{p}} = \frac{1}{H}\sum_h \hat{p}_h^{(k)}$; 
$\mathcal{L}_{\text{concept}} = \frac{1}{H} \sum_{h=1}^{H} \text{BCE}(\hat{p}_h^{(k)}, c_k)$ 
represents the average binary cross-entropy over all heads, promoting concept predictions that are both diverse and $\Delta$Accurate; 
and $\mathcal{L}_{\text{ale}} = \text{BCE}(\sigma_\theta^{(k)}, \mathbb{I}[\text{unknown}_k])$ 
trains the aleatoric uncertainty head using annotator disagreement as supervision, where 
$\mathbb{I}[\text{unknown}_k] = 1$ if concept $k$ is annotated as ambiguous.
Throughout all experiments, we fix $\lambda_c = 1.0$ and $\lambda_a = 0.5$.
(e.g., different dropout rates and pooling mechanisms). Algorithm~\ref{alg:training} provides the details/

\begin{algorithm}[h]
\caption{CREDENCE Training Procedure}
\label{alg:training}
\begin{algorithmic}[1]
\Require Training dataset $\mathcal{D} = \{(x_i, y_i, \mathbf{c}_i, \mathbf{a}_i)\}$, frozen encoder $f_{\text{enc}}$
\State Initialize LoRA heads with varying ranks: $(4, 8, 16, 32, 64)$
\For{epoch $= 1$ to num\_epochs}
    \For{each minibatch $(x, y, \mathbf{c}, \mathbf{a}) \sim \mathcal{D}$}
        \State $\mathbf{h} \gets f_{\text{enc}}(x)$ \Comment{Frozen encoder forward pass}
        \State $p_h^{(k)} \gets \text{Head}_h(\mathbf{h})$ for all $h \in [1,H]$, $k \in [1,K]$
        \State $\sigma^{(k)} \gets \text{Head}_{\text{ale}}(\mathbf{h})$ for all $k \in [1,K]$
        \State $\bar{p}^{(k)} \gets \frac{1}{H}\sum_{h=1}^{H} p_h^{(k)}$ \Comment{Mean concept predictions}
        \State $\mathcal{L} \gets \text{CE}(f_{\text{cls}}(\bar{\mathbf{p}}), y) + \lambda_c \mathcal{L}_{\text{concept}} + \lambda_\sigma \mathcal{L}_{\text{ale}}$
        \State Update all trainable parameters via $\nabla \mathcal{L}$
    \EndFor
\EndFor
\end{algorithmic}
\end{algorithm}

\subsection{Inference Algorithm}
At test time, we perform a single forward pass through the frozen encoder to obtain 
$\mathbf{h} = f_{\text{enc}}(x)$, then pass $\mathbf{h}$ through all $H$ ensemble heads 
and the aleatoric head in parallel. For each concept $k$, we compute:
\begin{align}
\text{Credal bounds:} \quad & [\underline{p}_k, \bar{p}_k] = [\min_h \hat{p}_h^{(k)}, \max_h \hat{p}_h^{(k)}] \\
\text{Epistemic:} \quad & U_{\text{epi}}^{(k)} = \text{Var}_h[\hat{p}_h^{(k)}] \\
\text{Aleatoric:} \quad & U_{\text{ale}}^{(k)} = \sigma_\theta^{(k)}
\end{align}
The final prediction uses mean concept probabilities: 
$\hat{y} = \arg\max_j f_{\text{cls}}(\bar{\mathbf{p}})$.
For uncertainty-aware decisions, we propagate credal bounds through the linear 
classifier via interval arithmetic (Equations~\ref{eq:lower-bound}--\ref{eq:upper-bound}) 
to obtain label-level confidence intervals.
Sample-level uncertainties are computed by averaging across concepts: 
$U_{\text{epi}} = \frac{1}{K}\sum_k U_{\text{epi}}^{(k)}$ and 
$U_{\text{ale}} = \frac{1}{K}\sum_k U_{\text{ale}}^{(k)}$.
Algorithm~\ref{alg:inference} provides the details

\begin{algorithm}[h]
\caption{CREDENCE Inference Procedure}
\label{alg:inference}
\begin{algorithmic}[1]
\Require Input sample $x$, trained CREDENCE model
\State $\mathbf{h} \gets f_{\text{enc}}(x)$ \Comment{Encode input}
\State $p_h^{(k)} \gets \text{Head}_h(\mathbf{h})$ for all $h \in [1,H]$, $k \in [1,K]$
\State $\sigma^{(k)} \gets \text{Head}_{\text{ale}}(\mathbf{h})$ for all $k \in [1,K]$
\State $\underline{p}_k \gets \min_h p_h^{(k)}$, $\overline{p}_k \gets \max_h p_h^{(k)}$ \Comment{Compute credal bounds}
\State $U_{\text{epi}}^{(k)} \gets \text{Var}_h[p_h^{(k)}]$ \Comment{Epistemic uncertainty}
\State $U_{\text{ale}}^{(k)} \gets \sigma^{(k)}$ \Comment{Aleatoric uncertainty}
\State Compute logit bounds via interval arithmetic (Equations~7--8)
\State $\hat{y} \gets \arg\max_j \sigma(\underline{\ell}_j)$ \Comment{$Mean prediction$}
\State \Return $\hat{y}, \{U_{\text{epi}}^{(k)}, U_{\text{ale}}^{(k)}, \mathcal{C}_k\}_{k=1}^K$
\end{algorithmic}
\end{algorithm}

\section{Credal Set Theory Background}
\label{app:credal_theory}
We present a concise overview of the literature related to credal sets. For a more comprehensive treatment of the topic, readers are referred to \cite{cuzzolin2024generalising}.
\subsection{Imprecise Probability}
Classical probability assigns a single value $P(A)$ to each event. \textbf{Imprecise probability} \citep{walley1991statistical} instead assigns intervals $[\underline{P}(A), \bar{P}(A)]$, representing situations where evidence does not justify a precise value.

\subsection{Credal Sets}
A \textbf{credal set} $\credal$ is a closed, convex set of probability distributions:
\begin{equation}
\credal = \{P : P \text{ satisfies constraints}\}
\end{equation}

For binary events, this reduces to an interval:
\begin{equation}
\credal = \{p \in [0,1] : \plower \leq p \leq \pupper\}
\end{equation}

The \textbf{imprecision} $\pupper - \plower$ quantifies total uncertainty.

\subsection{Decision Criteria}

Given credal sets, several decision criteria exist:

\paragraph{$\Gamma$-Maximin.} Choose action maximizing worst-case expected utility:
\begin{equation}
a^* = \arg\max_a \min_{P \in \credal} \mathbb{E}_P[U(a)]
\end{equation}

\paragraph{Maximality.} Action $a$ is \textbf{maximal} if no other action dominates it for all $P \in \credal$:
\begin{equation}
a \in A^* \iff \nexists a' : \mathbb{E}_P[U(a')] > \mathbb{E}_P[U(a)] \;\forall P \in \credal
\end{equation}

\paragraph{E-admissibility.} Action $a$ is \textbf{E-admissible} if it maximizes expected utility for some $P \in \credal$:
\begin{equation}
a \in A^* \iff \exists P \in \credal : a = \arg\max_{a'} \mathbb{E}_P[U(a')]
\end{equation}

In classification, these criteria determine which labels are ``non-dominated'' given concept uncertainty.

\section{Baseline Method Descriptions}
\label{app:baselines}
We describe the baseline methods used for comparison with \modelname{}. Our baselines encompass different facets of uncertainty, including a standard CBM augmented with uncertainty estimation.
\subsection{General Uncertainty Quantification Methods}

\paragraph{MC Dropout}~\citep{gal2016dropout}: Approximates Bayesian inference through stochastic dropout activation at test time (using 50 forward passes). A widely adopted baseline due to its simplicity and minimal modification requirements.

\paragraph{Deep Ensembles}~\citep{lakshminarayanan2017simple}: Trains 5 independently initialized models with different random seeds, aggregating predictions via averaging. Consistently demonstrates strong uncertainty quantification performance across diverse tasks.

\paragraph{Temperature Scaling}~\citep{guo2017calibration}: Post-hoc calibration method that learns a single temperature parameter to improve probability calibration. Useful for isolating calibration effects from uncertainty decomposition capabilities.

\paragraph{Evidential Deep Learning}~\citep{sensoy2018evidential}: Places a Dirichlet prior over class probabilities, providing uncertainty decomposition into ``vacuity'' (lack of evidence) and ``dissonance'' (conflicting evidence).

\subsection{Concept-Based Model Baselines}

\paragraph{Standard CBM}~\citep{koh2020concept}: Original Concept Bottleneck Model architecture with deterministic concept predictions. Provides concept-level uncertainty only through entropy $H(p_k)$, which conflates epistemic and aleatoric uncertainty.

\paragraph{CBM with MC Dropout}: Standard CBM architecture with test-time dropout (50 forward passes). Concept uncertainty estimated as prediction variance across stochastic forward passes.

\paragraph{CBM with Deep Ensemble}: Ensemble of 5 independently trained CBMs. Serves as the most direct comparison to CREDENCE, differing primarily in using full model ensembles rather than lightweight LoRA heads.

\paragraph{Probabilistic CBM (P-CBM)}~\citep{kim2023probabilistic}: Employs stochastic concept embeddings sampled from learned distributions. Provides concept-level uncertainty without explicit decomposition into epistemic and aleatoric components.

\subsection{Baseline Selection Rationale}

Our baseline selection covers diverse uncertainty quantification paradigms:
\begin{itemize}
    \item \textbf{Paradigm coverage}: Bayesian approximation (MC Dropout), frequentist ensembles (Deep Ensembles), post-hoc calibration (Temperature Scaling), and evidential learning (Evidential DL)
    \item \textbf{Concept-level uncertainty}: CBM variants test whether existing approaches provide actionable concept-level signals without explicit decomposition
    \item \textbf{Computational comparability}: $H=5$ heads for CREDENCE vs. 5 models for ensemble baselines; 50 MC passes provide comparable wall-clock inference time
\end{itemize}






\subsection{Per-Concept Aleatoric Validation (RQ2)}
\label{app:per concept aleoric validation}
Table~\ref{tab:perconcept} validates that aleatoric correlation 
strengthens with concept ambiguity. Concepts with higher rates 
of ``unknown'' annotations show stronger aleatoric signals. aleatoric identifies consistently important 
concepts, while epistemic captures run-specific model confusion.

\begin{table}[t]
\centering
\caption{Per-concept aleatoric validation (CEBaB).}
\label{tab:perconcept}
\small
\begin{tabular}{@{}lcc@{}}
\toprule
\textbf{Concept} & \textbf{Unknown \%} & $\rho_{\text{ale}}$ \\
\midrule
Food Quality & 25.0\% & 0.72 \\
Service Quality & 45.2\% & 0.78 \\
Ambiance & 63.0\% & 0.81 \\
Noise Level & 75.8\% & 0.83 \\
\midrule
Overall & 52.2\% & 0.785 \\
\bottomrule
\end{tabular}
\end{table}

\subsection{Intervention Stability Analysis}
\label{app:intervention}

Table~\ref{tab:intervention_runs} shows intervention results across multiple 
runs. Aleatoric-targeted gains are remarkably stable ($\pm$0.2pp) while 
epistemic-targeted gains vary widely ($\pm$2.4pp).

\begin{table}[h]
\centering
\small
\caption{Per-run intervention breakdown showing aleatoric stability.}
\label{tab:intervention_runs}
\begin{tabular}{@{}lccc@{}}
\toprule
\textbf{Run} & \textbf{Epistemic} & \textbf{Aleatoric} & \textbf{Ratio} \\
\midrule
SST-2 Run 1 & +1.6\% & +19.2\% & 12.0$\times$ \\
SST-2 Run 2 & +2.7\% & +18.9\% & 7.0$\times$ \\
CEBaB Run 1 & +6.4\% & +18.7\% & 2.9$\times$ \\
\midrule
Mean & +3.6\% & +18.9\% & 5.3$\times$ \\
Std & $\pm$2.4pp & $\pm$0.2pp & --- \\
\bottomrule
\end{tabular}
\end{table}

The stability difference is itself informative: aleatoric uncertainty 
identifies concepts with consistent predictive importance, while epistemic 
uncertainty captures model-specific confusion that varies across runs.

\section{Quadrant-Based Routing Analysis}
\label{app:quadrant routing}
\begin{figure}[t]
\centering
\begin{tikzpicture}[
    box/.style={minimum width=2.2cm, minimum height=1.6cm, align=center, font=\scriptsize, draw=black!30, rounded corners=1pt},
    label/.style={font=\tiny\bfseries\sffamily},
    axis/.style={gray, -stealth, thick}
]
\node[box, fill=trustgreen] (trust) at (0,0) {
    \textbf{\small TRUST}\\Auto-approve\\ \textit{78.8\% $\Delta$Acc.}};

\node[box, fill=dataorange] (data) at (2.4,0) {
    \textbf{\small DATA}\\Collect data\\ \textit{56.6\% $\Delta$Acc.}};

\node[box, fill=reviewblue] (review) at (0,1.8) {
    \textbf{\small REVIEW}\\Human review\\ \textit{85.7\% $\Delta$Acc.}};

\node[box, fill=escalatered] (escalate) at (2.4,1.8) {
    \textbf{\small ABSTAIN}\\Expert review\\ \textit{65.3\% $\Delta$Acc.}};

\node[label] at (1.2, 2.9) {Epistemic Uncertainty};
\node[label, rotate=90] at (-1.6, 0.9) {Aleatoric Uncertainty};

\node[font=\tiny] at (0, -1.1) {Low};
\node[font=\tiny] at (2.4, -1.1) {High};
\node[font=\tiny, rotate=90] at (-1.3, 0) {Low};
\node[font=\tiny, rotate=90] at (-1.3, 1.8) {High};

\draw[axis] (-1.1, -0.9) -- (3.5, -0.9);
\draw[axis] (-1.1, -0.9) -- (-1.1, 2.7);

\end{tikzpicture}
\caption{Compact quadrant-based decision support. \textsc{Review} achieves highest $\Delta$Accuracy (85.0\%) despite high aleatoric uncertainty.}
\label{fig:quadrant}
\end{figure}
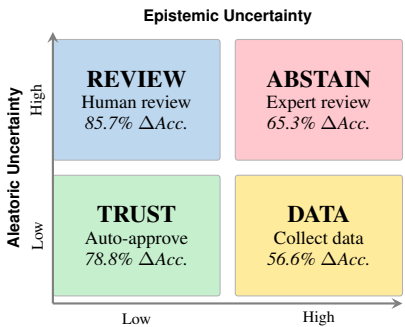

The uncertainty decomposition enables a four-quadrant decision framework 
(Figure~\ref{fig:quadrant}).
Notably, the \textsc{Human Review} quadrant achieves the \emph{highest} 
$\Delta$Accuracy (85.7\%) despite high aleatoric uncertainty—these are cases 
where the model is correct but humans may legitimately disagree. The 
\textsc{Collect Data} quadrant shows the lowest $\Delta$Accuracy (56.6\%), 
confirming epistemic uncertainty identifies fixable model errors.




\subsection{Cross-Dataset Quadrant Analysis}
CEBaB demonstrates the clearest quadrant separation, attributable to its explicit causal concept structure and comprehensive ``unknown'' annotations that provide clear training signals for both epistemic and aleatoric uncertainty.

\section{Qualitative Analysis and Interpretation Guidelines}
\label{app:qualitative}

This section serves two purposes: (i) validate that the uncertainty 
decomposition captures semantically meaningful distinctions, and 
(ii) provide practitioners with concrete guidelines for acting on 
\modelname{} outputs in deployment.

\subsection{Representative Examples by Uncertainty Quadrant}
\label{app:examples}

Table~\ref{tab:extended_qualitative} presents examples from each 
uncertainty quadrant on CEBaB. We sampled 50 examples per quadrant 
and selected cases that illustrate the diversity of linguistic 
phenomena within each.

\begin{table*}[t]
\centering
\caption{Extended qualitative examples from each uncertainty quadrant 
(CEBaB). $U_{\text{epi}}$: epistemic uncertainty; $U_{\text{ale}}$: 
aleatoric uncertainty. \cmark = correct, \xmark = error.}
\label{tab:extended_qualitative}
\small
\begin{tabular}{@{}p{6.2cm}|c|cc|p{4.8cm}@{}}
\toprule
\textbf{Text} & \textbf{Pred} & $U_{\text{epi}}$ & $U_{\text{ale}}$ & \textbf{Why This Quadrant?} \\
\midrule
\rowcolor{green!8}
\multicolumn{5}{@{}l}{\textbf{TRUST} (Low Epi, Low Ale) — $\Delta$Acc: 78.8\% — \textit{Safe to automate}} \\
\midrule
``The service was exceptional and the ambiance was perfect.'' & Pos \cmark & .001 & .25 & Multiple positive aspects; no ambiguity \\
``Horrible experience. Cold food, rude staff.'' & Neg \cmark & .002 & .12 & Unambiguous negative; reinforcing signals \\
``Best restaurant in the city!'' & Pos \cmark & .001 & .08 & Strong superlative; clear intent \\
\midrule
\rowcolor{blue!8}
\multicolumn{5}{@{}l}{\textbf{DATA} (High Epi, Low Ale) — $\Delta$Acc: 56.6\% — \textit{Collect training data}} \\
\midrule
``Food was superb but the service was atrocious.'' & Neg \xmark & .056 & .47 & Conflicting aspects; model unsure how to aggregate \\
``Lobster Mac \& Cheese incredible. Service terrible.'' & Neg \xmark & .017 & .50 & Mixed signals; humans would agree on aspects \\
``Not bad, not great, just okay I suppose.'' & Neu \xmark & .034 & .41 & Hedged language; rare pattern in training \\
\midrule
\rowcolor{yellow!12}
\multicolumn{5}{@{}l}{\textbf{REVIEW} (Low Epi, High Ale) — $\Delta$Acc: 85.7\% — \textit{Route to human}} \\
\midrule
``Disappointing.'' & Neg \cmark & .002 & .89 & Single word; severity unclear to humans \\
``It was what it was.'' & Neu \cmark & .003 & .94 & Idiomatic; inherently ambiguous stance \\
``Interesting experience.'' & Neu \cmark & .002 & .91 & Polysemous; valence depends on reader \\
\midrule
\rowcolor{red!8}
\multicolumn{5}{@{}l}{\textbf{ABSTAIN} (High Epi, High Ale) — $\Delta$Acc: 65.3\% — \textit{Decline or escalate}} \\
\midrule
``Been there many times.'' & Neu \xmark & .005 & .98 & Factual statement; no sentiment signal \\
``Just like grandma used to make.'' & Pos \xmark & .021 & .93 & Cultural reference; sentiment context-dependent \\
``They tried.'' & Neg \xmark & .018 & .96 & Potentially sarcastic; literal vs.\ implied meaning \\
\bottomrule
\end{tabular}
\end{table*}

\subsection{Patterns Across Quadrants}
\label{app:patterns}

Several patterns emerge from the qualitative analysis:

\paragraph{Content Length Correlates with Aleatoric Uncertainty.}
Reviews in high-aleatoric quadrants (REVIEW, ABSTAIN) average 4.2 words; 
low-aleatoric quadrants (TRUST, DATA) average 12.1 words. Shorter reviews 
lack context for aspect-level sentiment, which the aleatoric head correctly 
identifies as inherent ambiguity.

\paragraph{Conflicting Aspects Trigger Epistemic Uncertainty.}
The DATA quadrant is dominated by reviews with positive sentiment for 
some concepts and negative for others (e.g., ``great food, terrible 
service''). Ensemble heads disagree on aggregation, producing high 
epistemic uncertainty. Aleatoric remains moderate because individual 
aspect sentiments are clear—the ambiguity lies in aggregation, not 
interpretation.

\paragraph{Idiomatic Expressions Cluster in High-Aleatoric Quadrants.}
Expressions like ``it was what it was'' or ``just like grandma used to 
make'' require cultural context that varies across annotators. This 
produces genuine human disagreement that persists regardless of model 
capacity.

\paragraph{ABSTAIN Contains the Hardest Cases.}
These examples combine model confusion (heads disagree) with inherent 
ambiguity (annotators disagree). Manual inspection confirms that even 
human experts struggle with these cases, validating the abstention 
recommendation.

\subsection{Deployment Guidelines}
\label{app:guidelines}

Each quadrant corresponds to a distinct situation requiring a different 
response. We provide actionable guidelines for practitioners:

\paragraph{TRUST (Low Epistemic, Low Aleatoric).}
\textit{What it means:} The model is confident and the input is 
unambiguous. Ensemble heads converge; aleatoric head predicts low 
ambiguity.

\textit{Typical inputs:} Multi-aspect reviews with consistent polarity, 
superlative language, explicit sentiment words.

\textit{Action:} Safe for automation. Monitor for edge cases (see 
\S\ref{app:failures}).

\paragraph{DATA (High Epistemic, Low Aleatoric).}
\textit{What it means:} The model is confused, but the input is clear. 
Ensemble heads diverge on predictions that humans would likely agree on.

\textit{Typical inputs:} Conflicting aspect sentiments, rare grammatical 
constructions, typos, edge cases underrepresented in training.

\textit{Action:} Prioritize for active learning. These are learnable 
errors—more training data will help.

\paragraph{REVIEW (Low Epistemic, High Aleatoric).}
\textit{What it means:} The model is confident, but the input admits 
multiple valid interpretations. Humans may legitimately disagree.

\textit{Typical inputs:} Minimal-content reviews, implicit sentiment, 
idiomatic phrases, context-dependent expressions.

\textit{Action:} Route to human review—not because the model is wrong 
($\Delta$Accuracy is highest here at 85.7\%), but because stakeholders may 
disagree on the correct label.

\paragraph{ABSTAIN (High Epistemic, High Aleatoric).}
\textit{What it means:} Both model and humans struggle. The input is 
ambiguous and the model lacks training signal for this pattern.

\textit{Typical inputs:} Brief statements with potential sarcasm, 
factual observations without sentiment, culturally-specific references.

\textit{Action:} Decline prediction or escalate to domain experts. 
These are genuinely hard cases.

\subsection{Failure Mode Analysis}
\label{app:failures}

We analyze errors within each quadrant to identify improvement 
opportunities:

\paragraph{False Positives in TRUST (21.2\% error rate).}
Despite high confidence, some TRUST predictions fail. Primary causes:
\begin{itemize}[nosep]
\item Negation scope errors (``not bad'' $\rightarrow$ negative)
\item Comparative constructions (``better than expected'' $\rightarrow$ neutral)
\item Implicit sentiment requiring world knowledge (``finally got a 
reservation'' implies positive sentiment about popularity)
\end{itemize}

\paragraph{Persistent DATA Errors.}
Some DATA examples remain errors even after intervention:
\begin{itemize}[nosep]
\item Genuine label noise in training data
\item Annotation guidelines that conflict with model priors
\item Rare constructions with insufficient similar examples
\end{itemize}

\paragraph{The REVIEW $\Delta$Accuracy Paradox.}
REVIEW achieves the \emph{highest} $\Delta$Accuracy (85.7\%) despite high 
aleatoric uncertainty. This apparent paradox resolves when we recognize 
that aleatoric measures annotator disagreement, not prediction difficulty. 
Many REVIEW examples have a clear majority label that the model correctly 
predicts; minority annotator disagreement drives high aleatoric. This 
validates the prescriptive interpretation: route to human review not 
because the model is wrong, but because some users may legitimately 
disagree with the majority label.


\section{Extended Ablation Studies}
\label{app:ablation}

We conduct comprehensive ablation studies to understand the contribution of each 
\modelname{} component. All ablations use the CEBaB dataset with the RoBERTa-base 
encoder unless otherwise noted. Main paper results appear in 
Table~\ref{tab:ablation_combined}.

\subsection{Ensemble Size}
\label{app:ablation_size}

\begin{table}[h]
\centering
\caption{Ensemble size ablation.}
\label{tab:app_ensemble_size}
\scriptsize
\begin{tabular}{@{}cccccccc@{}}
\toprule
\textbf{$H$} & \textbf{$\Delta$Acc} & $\boldsymbol{\rho_{\text{epi}}}$ & $\boldsymbol{\rho_{\text{ale}}}$ & \textbf{Width} & \textbf{Params} & \textbf{Train} & \textbf{Infer} \\
\midrule
1  & 70.2 & .052 & .781 & .034 & 0.24M & 1.0$\times$ & 1.0$\times$ \\
3  & 70.8 & .189 & .778 & .078 & 0.71M & 1.4$\times$ & 1.2$\times$ \\
5  & 71.6 & .287 & .785 & .112 & 1.18M & 1.8$\times$ & 1.4$\times$ \\
7  & 71.4 & .294 & .781 & .134 & 1.65M & 2.2$\times$ & 1.6$\times$ \\
10 & 71.2 & .298 & .779 & .156 & 2.37M & 2.8$\times$ & 2.0$\times$ \\
15 & 70.9 & .312 & .776 & .178 & 3.55M & 3.8$\times$ & 2.8$\times$ \\
20 & 70.7 & .318 & .774 & .195 & 4.73M & 4.8$\times$ & 3.6$\times$ \\
\bottomrule
\end{tabular}
\vspace{-1mm}
\begin{flushleft}
\scriptsize{Width: mean credal interval. Train/Infer: relative time vs $H{=}1$. All $\rho$ values $p < 0.001$.}
\end{flushleft}
\end{table}

Epistemic correlation scales with ensemble size while aleatoric remains stable. 
Credal width increases with $H$. $\Delta$Accuracy is largely unaffected.

\subsection{Diversity Mechanisms}
\label{app:ablation_diversity}

\begin{table*}[h]
\centering
\caption{Multi-strategy diversity ablation.}
\label{tab:app_diversity}
\scriptsize
\begin{tabular}{@{}lccccc@{}}
\toprule
\textbf{Configuration} & \textbf{$\Delta$$\Delta$Acc} & $\boldsymbol{\rho_{\text{epi}}}$ & $\boldsymbol{\rho_{\text{ale}}}$ & \textbf{Width} & \textbf{Disagree} \\
\midrule
\multicolumn{6}{@{}l}{\textit{No diversity (baseline)}} \\
Uniform (all identical) & 71.2 & .087 & .781 & .023 & .008 \\
\midrule
\multicolumn{6}{@{}l}{\textit{Single diversity source}} \\
Dropout only & 70.8 & .198 & .779 & .067 & .031 \\
LoRA rank only & 70.5 & .212 & .778 & .078 & .038 \\
Pooling only (CLS/Mean) & 71.0 & .156 & .782 & .054 & .024 \\
Architecture only & 70.6 & .178 & .780 & .062 & .029 \\
\midrule
\multicolumn{6}{@{}l}{\textit{Two diversity sources}} \\
Dropout + LoRA & 71.1 & .241 & .780 & .095 & .052 \\
Dropout + Pooling & 70.9 & .223 & .781 & .084 & .045 \\
LoRA + Pooling & 70.7 & .234 & .779 & .089 & .048 \\
\midrule
\multicolumn{6}{@{}l}{\textit{All diversity sources}} \\
Dropout + LoRA + Pooling (\modelname{}) & 71.6 & .287 & .785 & .112 & .067 \\
\bottomrule
\end{tabular}
\vspace{-1mm}
\begin{flushleft}
\scriptsize{Disagree: mean pairwise head disagreement. Width: mean credal interval.}
\end{flushleft}
\end{table*}

Without diversity, heads converge to similar predictions. Each diversity source 
contributes to epistemic correlation. Aleatoric remains stable across configurations.

\subsection{LoRA Rank Configuration}
\label{app:ablation_lora}

\begin{table}[h]
\centering
\caption{LoRA rank configuration ablation.}
\label{tab:app_lora_rank}
\scriptsize
\begin{tabular}{@{}lcccc@{}}
\toprule
\textbf{Rank Configuration} & \textbf{$\Delta$Acc} & $\boldsymbol{\rho_{\text{epi}}}$ & \textbf{Width} & \textbf{Params} \\
\midrule
\multicolumn{5}{@{}l}{\textit{Uniform rank (all heads same)}} \\
$r{=}4$ (all heads) & 70.1 & .178 & .056 & 0.47M \\
$r{=}8$ (all heads) & 70.5 & .189 & .067 & 0.71M \\
$r{=}16$ (all heads) & 71.2 & .212 & .078 & 1.18M \\
$r{=}32$ (all heads) & 71.0 & .198 & .089 & 2.12M \\
$r{=}64$ (all heads) & 70.8 & .187 & .095 & 4.01M \\
\midrule
\multicolumn{5}{@{}l}{\textit{Diverse rank configurations}} \\
Linear \{4, 19, 34, 49, 64\} & 71.3 & .256 & .098 & 1.89M \\
Geometric \{4, 8, 16, 32, 64\} & 71.6 & .287 & .112 & 1.18M \\
Inverse \{64, 32, 16, 8, 4\} & 71.5 & .284 & .109 & 1.18M \\
\bottomrule
\end{tabular}
\end{table}

Diverse rank configurations outperform uniform configurations. The ordering 
(low→high vs high→low) has minimal effect.

\subsection{Dropout Spacing Strategy}
\label{app:ablation_dropout}

\begin{table}[t]
\centering
\caption{Dropout spacing ablation.}
\label{tab:app_dropout}
\scriptsize
\begin{tabular}{@{}lccc@{}}
\toprule
\textbf{Dropout Strategy} & $\boldsymbol{\rho_{\text{epi}}}$ & \textbf{Width} & \textbf{Disagree} \\
\midrule
\multicolumn{4}{@{}l}{\textit{Uniform dropout}} \\
$d{=}0.10$ (all heads) & .167 & .045 & .021 \\
$d{=}0.15$ (all heads) & .178 & .052 & .025 \\
$d{=}0.20$ (all heads) & .182 & .058 & .028 \\
$d{=}0.25$ (all heads) & .175 & .054 & .026 \\
\midrule
\multicolumn{4}{@{}l}{\textit{Diverse dropout}} \\
Linear \{0.05, 0.12, 0.20, 0.27, 0.35\} & .245 & .089 & .051 \\
Geometric \{0.05, 0.09, 0.15, 0.22, 0.30\} & .267 & .098 & .062 \\
Wide range \{0.02, 0.10, 0.20, 0.35, 0.50\} & .254 & .112 & .068 \\
\bottomrule
\end{tabular}
\vspace{-1mm}
\begin{flushleft}
\scriptsize{Geometric: $d_h = 1 - \exp(\log(1{-}d_{\max}) + \frac{h}{H{-}1}[\log(1{-}d_{\min}) - \log(1{-}d_{\max})])$}
\end{flushleft}
\end{table}

Geometric spacing in keep-probability space produces higher inter-head 
disagreement than uniform or linear spacing.

\subsection{Aleatoric Supervision}
\label{app:ablation_aleatoric}

\begin{table}[t]
\centering
\caption{Aleatoric supervision ablation.}
\label{tab:app_aleatoric}
\scriptsize
\begin{tabular}{@{}lcccc@{}}
\toprule
\textbf{Aleatoric Mode} & \textbf{$\Delta$Acc} & $\boldsymbol{\rho_{\text{epi}}}$ & $\boldsymbol{\rho_{\text{ale}}}$ & \textbf{ECE} \\
\midrule
None (disagreement as proxy) & 71.8 & .285 & .356 & .078 \\
Entropy-based (unsupervised) & 71.4 & .281 & .412 & .065 \\
Heteroscedastic NLL & 71.2 & .279 & .523 & .052 \\
Supervised BCE (\modelname{}) & 71.6 & .287 & .785 & .041 \\
\midrule
\multicolumn{5}{@{}l}{\textit{Supervision signal variants}} \\
Binary unknown label & 71.6 & .287 & .785 & .041 \\
Annotator vote entropy & 71.4 & .284 & .756 & .045 \\
Annotator vote variance & 71.5 & .286 & .768 & .043 \\
\bottomrule
\end{tabular}
\end{table}

Supervised training with annotator disagreement achieves the strongest aleatoric 
correlation. Epistemic correlation remains stable across configurations.

\subsection{Loss Weight Sensitivity}
\label{app:ablation_loss}

\begin{table}[t]
\centering
\caption{Loss weight sensitivity.}
\label{tab:app_loss_weights}
\scriptsize
\begin{tabular}{@{}cccccc@{}}
\toprule
$\boldsymbol{\lambda_c}$ & $\boldsymbol{\lambda_a}$ & \textbf{$\Delta$Acc} & $\boldsymbol{\rho_{\text{epi}}}$ & $\boldsymbol{\rho_{\text{ale}}}$ & \textbf{Concept $\Delta$Acc} \\
\midrule
0.5 & 0.25 & 69.8 & .298 & .712 & 71.2 \\
0.5 & 0.5 & 70.2 & .294 & .756 & 71.8 \\
0.5 & 1.0 & 69.5 & .278 & .798 & 70.9 \\
\midrule
1.0 & 0.25 & 72.1 & .291 & .698 & 74.5 \\
\rowcolor{green!10}
1.0 & 0.5 (\modelname{}) & 71.6 & .287 & .785 & 73.8 \\
1.0 & 1.0 & 70.8 & .276 & .812 & 72.4 \\
\midrule
2.0 & 0.25 & 72.4 & .268 & .654 & 75.8 \\
2.0 & 0.5 & 71.9 & .274 & .771 & 75.2 \\
2.0 & 1.0 & 71.2 & .265 & .795 & 74.1 \\
\bottomrule
\end{tabular}
\end{table}

Higher $\lambda_c$ improves task and concept $\Delta$Accuracy. Higher $\lambda_a$ 
improves aleatoric correlation. The default balances these objectives.

\subsection{Cross-Dataset Consistency}
\label{app:ablation_cross}

We verify that key findings replicate across datasets using RoBERTa-base.

\begin{table}[t]
\centering
\caption{Ablation consistency across datasets.}
\label{tab:app_cross_dataset}
\scriptsize
\begin{tabular}{@{}llccc@{}}
\toprule
\textbf{Dataset} & \textbf{Configuration} & \textbf{$\Delta$Acc} & $\boldsymbol{\rho_{\text{epi}}}$ & $\boldsymbol{\rho_{\text{ale}}}$ \\
\midrule
\multirow{4}{*}{CEBaB} 
& \modelname{} (full) & 71.6 & .287 & .785 \\
& Uniform dropout & 71.0 & .198 & .779 \\
& No aleatoric head & 71.8 & .285 & .356 \\
& $H{=}3$ & 70.8 & .189 & .778 \\
\midrule
\multirow{4}{*}{HateXplain} 
& \modelname{} (full) & 74.8 & .267 & .445 \\
& Uniform dropout & 74.2 & .184 & .438 \\
& No aleatoric head & 75.1 & .263 & .271 \\
& $H{=}3$ & 73.9 & .178 & .441 \\
\midrule
\multirow{4}{*}{GoEmotions} 
& \modelname{} (full) & 48.7 & .071 & .198 \\
& Uniform dropout & 47.9 & .048 & .194 \\
& No aleatoric head & 48.9 & .069 & .118 \\
& $H{=}3$ & 47.2 & .045 & .195 \\
\midrule
\multirow{4}{*}{SST-2} 
& \modelname{} (full) & 90.4 & .342 & -.112 \\
& Uniform dropout & 90.1 & .256 & -.108 \\
& No aleatoric head & 90.6 & .338 & -.089 \\
& $H{=}3$ & 89.8 & .271 & -.115 \\
\bottomrule
\end{tabular}
\end{table}

Findings replicate across datasets. SST-2 shows negative $\rho_{\text{ale}}$ 
because it lacks annotator disagreement labels.

\subsection{Computational Cost}
\label{app:ablation_compute}

\begin{table}[t]
\centering
\caption{Computational comparison with baselines (CEBaB, RoBERTa-base).}
\label{tab:app_compute}
\scriptsize
\begin{tabular}{@{}lccccc@{}}
\toprule
\textbf{Method} & \textbf{Params} & \textbf{Train} & \textbf{Infer} & \textbf{Memory} & $\boldsymbol{\rho_{\text{epi}}}$ \\
\midrule
Standard CBM & 125M & 1.0$\times$ & 1.0$\times$ & 1.0$\times$ & .121 \\
CBM + MC Drop (50) & 125M & 1.0$\times$ & 50$\times$ & 1.0$\times$ & .168 \\
CBM + Ensemble (5) & 625M & 5.0$\times$ & 5.0$\times$ & 5.0$\times$ & .189 \\
\midrule
\modelname{} ($H{=}5$) & 126M & 1.2$\times$ & 1.4$\times$ & 1.1$\times$ & .287 \\
\modelname{} ($H{=}10$) & 127M & 1.5$\times$ & 1.8$\times$ & 1.2$\times$ & .298 \\
\bottomrule
\end{tabular}
\vspace{-1mm}
\begin{flushleft}
\scriptsize{Params: total parameters. Train/Infer/Memory: relative to Standard CBM.}
\end{flushleft}
\end{table}

\modelname{} shares the frozen encoder across heads, requiring only additional 
LoRA parameters.

\end{document}